%% file: main.tex
\documentclass{article} 
\usepackage{collas2025_conference,times}
\usepackage{easyReview}

\input{math_commands.tex}

\usepackage{hyperref}
\hypersetup{
    colorlinks=true,
    linkcolor=red,
    filecolor=magenta,
    urlcolor=blue,
    citecolor=purple,
    pdftitle={Overleaf Example},
    pdfpagemode=FullScreen,
    }

\usepackage[capitalize,sort&compress,nameinlink]{cleveref}
\crefname{section}{Sec.}{Secs.}
\Crefname{section}{Section}{Sections}
\Crefname{table}{Table}{Tables}
\crefname{table}{Tab.}{Tabs.}

\usepackage{adjustbox}
\usepackage{multirow}
\usepackage{booktabs}
\newcommand{\boldhline}{\specialrule{0.15em}{0em}{0.1em}}
\renewcommand{\arraystretch}{1.25}
\usepackage[tableposition=above]{caption}
\usepackage{enumitem}

\usepackage{xspace}

\makeatletter
\DeclareRobustCommand\onedot{\futurelet\@let@token\@onedot}
\def\@onedot{\ifx\@let@token.\else.\null\fi\xspace}

\def\eg{\emph{e.g}\onedot} 
\def\ie{\emph{i.e}\onedot} \def\Ie{\emph{I.e}\onedot}
 
 \def\vs{\emph{vs}\onedot}

\makeatother

\newcommand{\forgetscore}[1]{\textcolor{red}{-#1}}

\title{CLoRA: Parameter-Efficient Continual Learning\\with Low-Rank Adaptation}

\author{
	Shishir Muralidhara$^1$ \qquad Didier Stricker$^{1,2}$ \qquad René Schuster$^{1,2}$  \\
	$^1$Augmented Vision Group, German Research Center for Artificial Intelligence (DFKI) \\
	$^2$RPTU -- University of Kaiserslautern-Landau, Kaiserslautern \\
	{\texttt{firstname.lastname@dfki.de}}
}

\preprintcopy 

\begin{document}

\maketitle

\begin{abstract}
In the past, continual learning (CL) was mostly concerned with the problem of catastrophic forgetting in neural networks, that arises when incrementally learning a sequence of tasks.
Current CL methods function within the confines of limited data access, without any restrictions imposed on computational resources. 
However, in real-world scenarios, the latter takes precedence as deployed systems are often computationally constrained. 
A major drawback of most CL methods is the need to retrain the entire model for each new task. 
The computational demands of retraining large models can be prohibitive, limiting the applicability of CL in environments with limited resources.
Through CLoRA, we explore the applicability of Low-Rank Adaptation (LoRA), a parameter-efficient fine-tuning method for class-incremental semantic segmentation.
CLoRA leverages a small set of parameters of the model and uses the same set for learning across all tasks. 
Results demonstrate the efficacy of CLoRA, achieving performance on par with and exceeding the baseline methods.
We further evaluate CLoRA using NetScore, underscoring the need to factor in resource efficiency and evaluate CL methods beyond task performance.
CLoRA significantly reduces the hardware requirements for training, making it well-suited for CL in resource-constrained environments after deployment.
\end{abstract}

\section{Introduction}
\label{sec:intro}
While neural networks have demonstrated remarkable performance in deep learning across various domains, their rigid structure is prohibitive in adapting to new tasks.
Typically, neural networks are trained on a fixed dataset, but real-world scenarios are non-stationary \citep{embracingChange} where data drift occurs, which can impact model performance \citep{nonstationaryLearning}, and objectives may change over time, such as the introduction of new classes or tasks \citep{embracingChange}.
Fine-tuning or transfer learning on the new data would lead to catastrophic forgetting \citep{catastrophicForgetting}, causing the network to perform poorly on the previously learned tasks, this stems from the inherent plasticity of neural networks where previously learned information is overwritten during the learning of new tasks. 
Incrementally learning results in the stability-plasticity dilemma \citep{stabilityPlasticity}, where networks with higher stability are restrictive in learning new tasks, and higher plasticity tend to overwrite previously learned information.
A straightforward solution to prevent forgetting is to retrain the model with all encountered data, but this requires significant computational resources \citep{greenCL}, training time, storage, and may not always be feasible due to data unavailability \citep{clRoboticsSurvey}.
\begin{figure}[t]
	\centering
	\includegraphics[width=0.5\columnwidth]{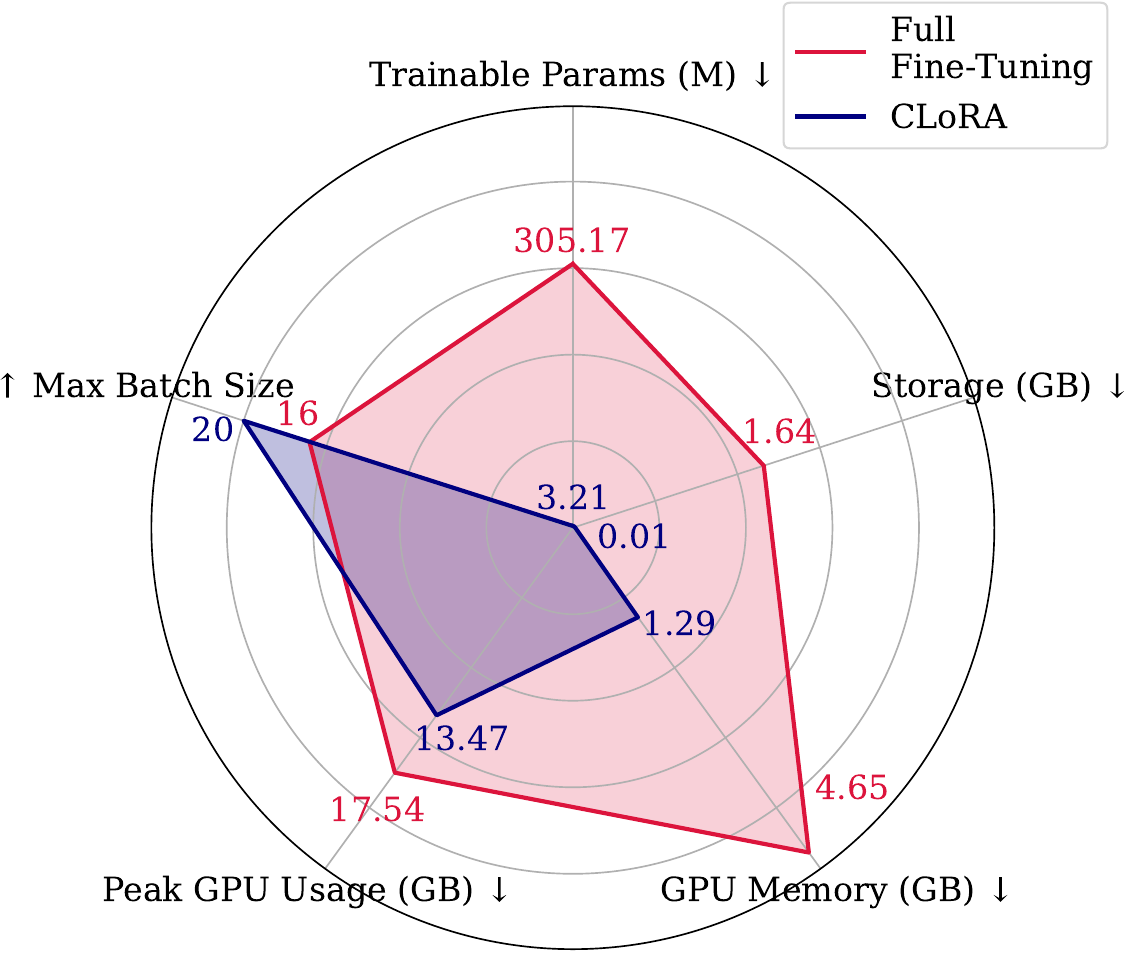}
	\caption{Comparison of resource efficiency of CLoRA against full fine-tuning, under identical conditions. }
	\label{fig:clora_fft_comparison}
\end{figure}
In contrast to traditional machine learning, where models are trained in isolation and fixed, continual learning (CL) is a dynamic learning paradigm that more accurately mirrors the non-stationary nature of the real world.
Continual learning involves incrementally learning a sequence of tasks \citep{clScenarios}, while minimizing catastrophic forgetting on previous tasks under restricted or no access to previously encountered data.
Each task can represent a change either in the input or output distribution resulting in two main incremental-learning settings. 
In domain-incremental learning, a task represents a change in the input distribution, such as different image sources. 
This can also be extended to learning modalities in the case of modality-incremental learning \citep{MIL}.
Class-incremental learning represents a shift in the output distribution by incorporating novel, previously unseen classes, and extends to learning evolving classes in CLEO \citep{CLEO}.
CL therefore focuses on efficiently adapting an existing model to new tasks, circumventing the need for retraining from scratch. 
However, most CL methods other than those based on transfer-learning are designed under the assumption of having access to offline resources \citep{budgetedCL}, overlooking computational constraints of deployed systems which could be prohibitive in updating large networks. 
In this work, we address both storage and computational constraints and present an approach that adheres to the original constraint of restricted data access, while being able to operate under resource constrained environments, without compromising on the choice of networks.  

Parameter-efficient fine-tuning (PEFT) methods were developed for adapting Large Language Models (LLMs) to specific downstream tasks \citep{peftNLP}.
PEFT methods significantly reduce the number of trainable parameters while still achieving performance on-par with full fine-tuning. 
PEFT offers several advantages over full fine-tuning, such as reducing hardware requirements, and avoiding overfitting on the fine-tuned dataset \citep{peftSurvey}.
PEFT with CL (PECL) minimizes the number of trainable parameters updated during adaptation, thus allowing for more efficient usage of computational resources, which is crucial in scenarios where resources are limited.
\citep{cssSurvey} note that within current CL methods for semantic segmentation, those utilizing stronger backbones consistently achieve superior performance across both old and new classes. PECL facilitates leveraging the capabilities of extremely large models, even within constrained environments.
In this work, we introduce Continual Learning with Low Rank-Adaptation (CLoRA), the first PECL method that uses LoRA \citep{LoRA} for class-incremental semantic segmentation, which forms our primary contribution.
A key strength of CLoRA lies in its modular compatibility, since it is agnostic to the regularization strategy used to mitigate forgetting, making it a lightweight, resource-efficient extension to existing CL methods.
We validate this versatility through extensive experiments using several baselines such as MiB \citep{MiB}, RCIL \citep{RCIL}, SATS \citep{SATS}, and SSUL \citep{SSUL} across different segmentation networks.
CLoRA offers resource efficiency advantages over traditional CL, as illustrated in \cref{fig:clora_fft_comparison} and detailed under \cref{sec:clora_fft}, in which we show that CLoRA requires less hardware resources while achieving comparable or superior segmentation results to our baselines.

\section{Background}
\label{sec:background}
Continual learning encompasses methods for enabling a system to incrementally learn new information, while retaining previously learned knowledge. 
These methods can be categorized into three main categories: Architecture-, replay-, and regularization-based approaches.
Additionally, there are hybrid approaches that integrate a combination of these methods.
A more detailed review and survey of these approaches is presented by \cite{cl_comprehensive_survey}.

\subsection{Continual Semantic Segmentation}
\label{sec:css}
MiB \citep{MiB} proposes a novel distillation loss to account for background shift in incremental segmentation by comparing the background class prediction by the old model with background and new class predictions by the new model.
PLOP \citep{PLOP} addresses the background shift by using pseudo-labels for the background pixels predicted by the previous task model. 
SATS \citep{SATS} uses self-attention maps from transformers and class-specific region pooling is used for between and within class knowledge distillation.
RCIL \citep{RCIL} maintains two branches during training, one branch is frozen after initial training and preserves the knowledge of old classes, whereas the other branch is trainable and is used for learning new tasks. 
REMINDER \citep{REMINDER} uses a class-similarity based distillation, to distill knowledge from a previous model with classes similar to the new classes.
AWT \citep{AWT} addresses background shift through classifier initialization by identifying the most relevant weights from the previous background for the new classes.
EWF \citep{EWF} merges the trained models of previous and current tasks, weighted by a merging factor. 
SSUL-M \citep{SSUL} addresses background shift by introducing an unknown class that separates the future classes from the background and uses pseudo-labeling for previous classes. 
ALIFE \citep{ALIFE} stores feature representations extracted from previous models for replay instead of explicitly storing images from previous tasks.
RECALL \citep{RECALL} uses a generic generative model or web-crawler for generating/retrieving images of previous classes and pseudo-labeling. 
DiffusePast \citep{DiffusePast} addresses issues with GAN-generated images, and uses a stable diffusion model for generating accurate images. 
In this work, we focus on class-incremental learning with knowledge distillation. 
\subsection{Transfer-Learning-based Continual Learning}
Transfer-learning-based CL methods leverage large pretrained networks for feature extraction and only train the classifier for incrementally learning new classes.  
Typically, these methods use a pretrained network, that is either frozen directly or after training for the initial task. 
By forgoing full network retraining, this approach addresses the computational constraints in continual learning.
\cite{off-the-shelf-CL} proposes a straightforward approach that extracts features and stores class prototypes for a prototype-based classification.
FeTrIL \citep{FeTrIL} uses a pseudo feature generator to represent past classes through geometric translations of new class features and old class prototypes. A linear classifier is then jointly trained on new classes and old classes. 
Adapt and Merge (APER) \citep{APER} adapts the pretrained model to incremental data using PEFT to bridge the domain gap.
Subsequently, it aggregates the adapted and pretrained model embeddings and freezes them for incremental tasks.
RanPAC \citep{RanPac} uses random projection layers to map the features extracted from the pretrained network into a higher-dimensional space, increasing class separation.
FeCAM \citep{FeCAM} highlights the shortcomings of Euclidean distance based prototype classification in incremental learning and proposes using a Bayesian classifier.

\subsection{Parameter Efficient Continual Learning}
\label{sec:pecl}
There are several parameter-efficient fine-tuning (PEFT) methods \citep{peftSurvey} like: Additive fine-tuning, which introduces additional parameters through adapters \citep{adapterDrop, IA3} or prompts \citep{promptTuning, prefixTuning}; partial fine-tuning \citep{Bitfit, peftMasking, peftPruning} where a subset of pretrained parameters are selected; reparametrized  fine-tuning \citep{LoRA, DyLoRA, QLoRA} which uses low-rank transformation to reduce the number of trainable parameters.
The utilization of PEFT in continual learning is gaining traction, leading to the development of computationally efficient approaches, we refer to as parameter-efficient continual learning (PECL).
\cite{IRU} propose a dynamically growing network with incremental rank updates, where for each task, a new trainable rank-1 matrix is added while the previous low-rank matrices are frozen. 
During inference, it requires the task-ID for selecting the appropriate weights. 
Continual learning with low rank adaptation (CoLoR) \citep{CoLoR} uses LoRA for training expert models and k-means clustering for storing k cluster centers for each dataset. 
During inference, the task-ID is inferred by determining the nearest cluster center and the corresponding expert model is selected. 
This additional step incurs an additional computational overhead during both training and inference. 
\cite{taskArithmeticLoRA} use LoRA to train expert models for each task, and then merges them using task arithmetic \citep{taskArithmetic}.  
It requires fine-tuning on a small subset of data gathered from each class across all tasks, similar to a rehearsal-based approach.
LAE \citep{LAE} framework consists of three stages: Learning new tasks by leveraging  pretrained models with an online PEFT module, accumulating task-specific knowledge into an offline PEFT module, and ensembling during inference using the online and offline modules.
Orthogonal low-rank adaptation \citep{OLoRA} incrementally adds LoRA for each task and ensures orthogonality between the current and previous modules to mitigate interference between tasks and minimize forgetting.
We present a PECL method, that leverages a single LoRA module for learning across all tasks. 
We discuss the challenges of using task-specific modules for segmentation in \cref{sec:ciss_challenges}.

\section{CLoRA: Continual Low-Rank Adaptation}
\label{sec:clora}

Continual learning involves learning a sequence of tasks $T= \{t_0, t_1,..., t_n \}$, where each task is associated with task-specific data $(X_t, Y_t)$.
Depending upon the incremental setting, either the distribution of $X_t$ or $Y_t$ varies across tasks.
In class-incremental learning, the input distribution remains consistent and in each task, subsets $C_t \subset C$ of non-overlapping classes $C_i \cap C_{j} = \emptyset, i \neq j$ are introduced.
These subsets compose the totality of classes $C = C_0 \cup C_1 \cup ... \cup C_t$.


\subsection{Low-Rank Adaptation (LoRA)}
\label{sec:lora}
Low-rank adaptation (LoRA) \citep{LoRA} is a parameter-efficient fine-tuning method that uses reparameterization for adapting pretrained models to downstream tasks.
LoRA uses low-rank transformation to significantly reduce the number of trainable parameters and the computational requirements while still achieving performance on par with full fine-tuning.
For a pretrained network with weights $W \in \mathbb{R}^{d \times k}$, LoRA fine-tunes a very small subset of weights $\triangle W$, represented using two low-rank matrices, $A \in \mathbb{R}^{d \times r} $ and $B \in \mathbb{R}^{r \times k}$, where $r$ is the rank of the matrix and $r \ll min(d,k)$. $A$ is initialized using random Gaussian distribution, and $B$ is initialized as a zero matrix. The rank $r$ is a hyperparameter that determines the number of trainable parameters.
The two low-rank matrices are substituted in place of the original weights $W$ in each layer of a transformer, and only these low-rank matrices, which are typically a fraction of the original weights, are trainable. 
Furthermore, LoRA is applied only to the query and value projection layers, which contributes to further reducing the number of trainable parameters.
Additionally, LoRA also drastically reduces the storage footprint, allowing to store modules for each task and switching only the task-specific LoRA modules, while the pretrained weights remain constant. 
During training, the pretrained weights are frozen and only the LoRA weights are updated. The forward pass is modified and the output $h$ is calculated as 
\begin{equation}
	h = W(x) + \triangle W(x) = W(x) + BA(x)
\end{equation}
Unlike other PEFT methods such as adapters, which add inference overhead, LoRA avoids inference latency by merging the LoRA modules with the pretrained weights. The weights are updated as $W^{'} = W +BA$.
\begin{figure}[t]
	\centering
	\setlength{\tabcolsep}{1pt}
	
	\begin{tabular}{ccccc}
		Image & Ground Truth & Task 0 & Task 1  \\ 
		
		\includegraphics[width=.2\linewidth]{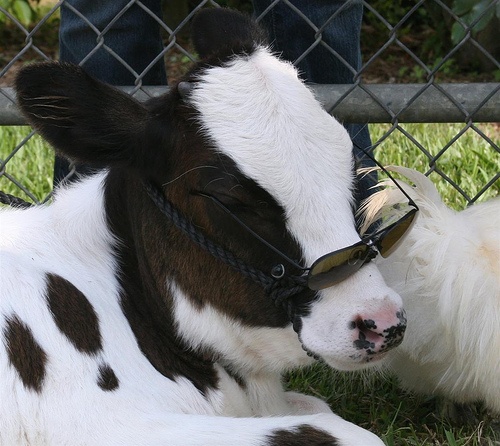} &
		\includegraphics[width=.2\linewidth]{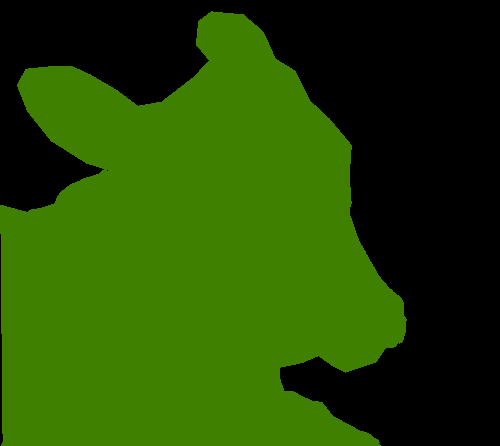} &
		\includegraphics[width=.2\linewidth]{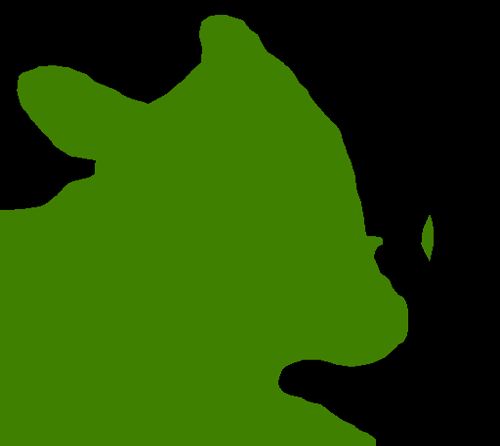} &
		\includegraphics[width=.2\linewidth]{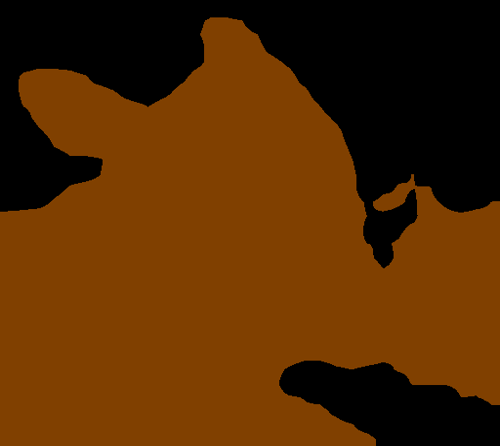} 
		
	\end{tabular}
	\caption{Conflicting predictions from task-specific modules on the PASCAL VOC \citep{PASCAL} dataset using the \textit{15-5} setting, in which the two modules have conflicting predictions.}
	\label{fig:conflicts}
\end{figure}

\subsection{Challenges in Class-Incremental Semantic Segmentation}
\label{sec:ciss_challenges}

Current PECL methods for image classification \citep{IRU, CoLoR} mostly leverage individual, task-specific LoRA modules for incrementally learning new tasks. 
This approach holds significant appeal, achieving performance on par with full fine-tuning while entirely mitigating catastrophic forgetting as it avoids overwriting of information. 
Furthermore, it demonstrates storage efficiency compared to methods using dynamically expanding networks \citep{progressiveNNs}, with LoRA modules consuming only a fraction of the memory required for storing complete models.
While these factors make using task-specific LoRA modules a compelling choice, its application to class-incremental \textit{semantic segmentation} presents several challenges.
Unlike class-incremental classification, where images typically have a single label, and the task-ID can be used to select the expert model trained on that specific class for inference, task-ID inference in class-incremental segmentation is not straightforward as the tasks are not mutually exclusive.
In segmentation, images are typically annotated with multiple classes, which may span across different tasks in a class-incremental setting, and the final prediction may require combining predictions of classes learned across different tasks, with task-specific LoRA modules.
A potential solution is to use all task-specific modules during inference and merge their results (see appendix for details). 
However, with this approach the inference time increases proportionally with number of tasks, making it ineffective.
Furthermore, merging task-wise predictions is particularly challenging due to \textit{background shift} in incremental segmentation, where the background class is a catch-all class that encompasses all previously seen and potential future classes.
During incremental learning, only the current task classes are annotated, and the remaining classes are labeled as the background class. This results in the task-specific modules being unaware of past and future classes. 
Consequently, the definition of the background changes across tasks, leading to inconsistent predictions of the background by the individual modules. 
Due to the isolated nature of task-specific modules, they are not aware of classes learned in other tasks through different modules. 
As a consequence, they make predictions based only on the subset of classes they have encountered.
This can lead to conflicting predictions for visually similar classes, with different modules predicting different classes for the same pixel.
Resolving these discrepancies and determining the correct prediction poses a significant challenge.

\Cref{fig:conflicts} illustrates this challenge using the \textit{15-5} setting in PASCAL VOC. 
For the class \textit{cow} which was learned in task 0, the task 1 module mistakenly predicts the class as \textit{sheep}. 
This error arises because the two classes are visually similar, and during task 1, the module has not seen images of \textit{cow} to discriminate between \textit{cow} and \textit{sheep}.

\subsection{Proposed Approach}
\label{sec:proposed_approach}

\begin{figure*}[t]
\centering
\includegraphics[width=0.85\textwidth]{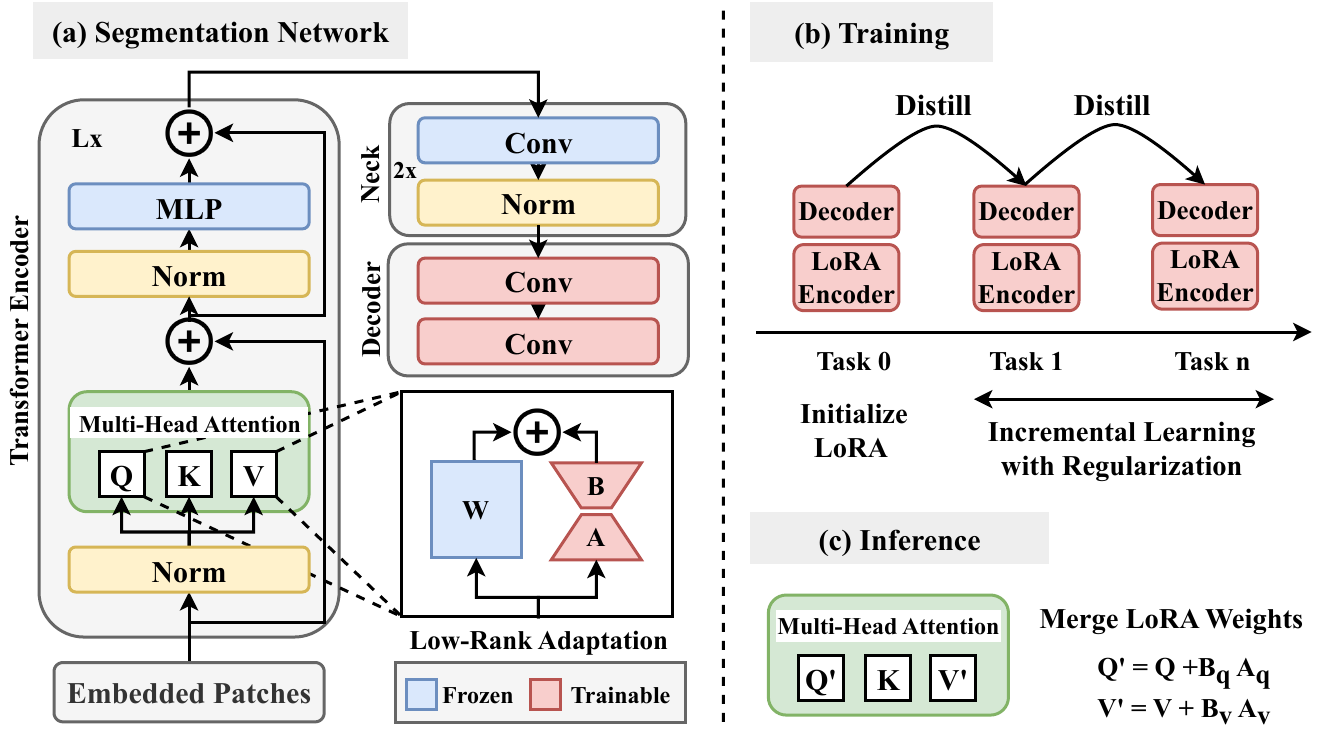}
\caption{Continual Learning with Low-Rank Adaptation (CLoRA). (a) CLoRA uses low-rank adaptation for parameter-efficient fine-tuning of Vision Transformers in resource-constrained environments. The decoder undergoes fine-tuning while all other weights remain frozen. 
	(b) Training: The encoder using LoRA is initialized for the initial task and trained. For the subsequent tasks, the same LoRA weights and decoder are updated using knowledge distillation.
	(c) Inference: After learning all tasks, the LoRA weights are merged with the frozen weights. This approach mitigates additional inference time and parameters and facilitates inference without task-ID.}
\label{fig:CLoRA}
\end{figure*}
Considering the limitations associated with dynamically growing task-specific modules for incremental learning in semantic segmentation, we introduce CLoRA. 
CLoRA leverages a single LoRA module to incrementally learn tasks with knowledge distillation. This approach addresses the aforementioned limitations and presents several advantages: 
Unlike methods that add new LoRA modules for each task, CLoRA maintains a consistent network architecture size throughout the learning process. 
Utilizing a single LoRA module across all tasks results in a task-agnostic model, circumventing the challenges related to task-ID inference and conflicting predictions typically associated with employing multiple LoRA modules. 
Additionally, as CLoRA does not necessitate inference and merging of multiple tasks, the inference time remains constant. 
\newline\newline
An overview of CLoRA is presented in \cref{fig:CLoRA}. 
For the first task, the LoRA module is initialized, and the encoder weights are substituted with LoRA, by freezing the encoder and training only the LoRA weights similar to PEFT. 
The decoder, which constitutes a very small portion of the network undergoes full fine-tuning. 
Subsequently, when new tasks are introduced incrementally, the same trained LoRA weights from the previous task are reused and updated. 
To preserve knowledge from previous tasks and mitigate catastrophic forgetting, distillation is used. Specifically, we utilize the knowledge distillation loss proposed by MiB \citep{MiB}. However, it is feasible to integrate other regularization approaches with minimal adjustments, as shown in our experiments (\cref{tab:pascal_segformer,sec:experiments:additional:extension}).

During incremental training, knowledge distillation loss is used to transfer knowledge from the previous task model, acting as the teacher $f_{t-1}$ to the student model $f_t$ being trained on the current task $t$. 
However, in class-incremental semantic segmentation, this approach faces the background shift problem.
For the current task $t$ with the set of classes  $C_t$,  the teacher model, trained on previous tasks, may have learned the current task classes as background in the earlier class set $C_{0:t-1}$. 
As a result, during distillation, there is a mismatch: the teacher predicts background class for pixels belonging to new classes, while the student correctly predicts them to the actual classes in $C_t$.
This mismatch affects the balance between stability and plasticity, limiting the model’s ability to adapt to new classes.
Addressing this limitation, MiB \citep{MiB} introduces a novel distillation strategy by aggregating the student logits for the new classes with the background logits before applying knowledge distillation.
\newline\newline
Through knowledge distillation, we can utilize the same LoRA weights across tasks, and  ensure the LoRA module is aware of classes learned across different tasks and avoids conflicting predictions.
Finally, after learning all tasks, the LoRA weights can be merged back into the encoder to update the original weights. 
This approach helps avoid additional inference overhead such as additional computation \citep{IRU}, determining task-ID and selecting task-experts \citep{CoLoR}, or merging of multiple task outputs.

\section{Experiments and Results}
\label{sec:results}
In this section we discuss the datasets used to evaluate CLoRA, implementation details, and comparison baselines.
We outline the task settings for each dataset and report their results (visualizations in the appendix). 
All reported results are evaluated in terms of mean Intersection over Union (mIoU).
To assess the extent of forgetting, we include the forget score (FS), by comparing the model's performance after learning all tasks with the corresponding joint training (JT) baseline, which is used to approximate the upper bound.
We highlight the efficacy of CLoRA against a frozen encoder, explore the influence of varying ranks, and reinitializing the LoRA module after each task. 
We demonstrate that CLoRA can be extended to integrate with other networks and approaches. Additionally, we quantify the resource efficiency of CLoRA using NetScore \citep{NetScore}.
\subsection{Datasets}
\label{sec:datasets}

\begin{itemize}[itemsep=6pt]
	
	\item \textbf{PASCAL VOC} part of the Visual Object Classes Challenge \citep{PASCAL} contains images and annotations for 21 classes such as animals, person, vehicles and household items, including the background. 
	
	\item \textbf{ADE20K} \citep{ADE} consists of over 25k images for training and 2k images for testing. 
	The dataset covers 150 classes, allowing to design CL tasks with sizable number of classes being added in each increment.	
	
	\item \textbf{Cityscapes} \citep{cityscapes} comprises urban environment images, with 2975 for training and 500 for validation. It includes dense annotations for 19 classes, posing challenges for segmentation tasks.

\end{itemize}

\begin{table*}[t]
	\caption{Results on PASCAL VOC \citep{PASCAL} dataset after learning all tasks.}
\begin{adjustbox}{width=\textwidth}
	\centering
	\begin{tabular}{c||c|c|c|c||c|c|c|c||c|c|c|c||c|c|c|c}
		\boldhline
		\multirow{2}{*}{\textbf{Method}} & \multicolumn{4}{c||}{\textbf{15-5}} & \multicolumn{4}{c||}{\textbf{15-1}} & \multicolumn{4}{c||}{\textbf{5-3}} & \multicolumn{4}{c}{\textbf{10-1}} \\ \cline{2-17}
		&  0-15  &  16-20 &  \textbf{All}  & \textbf{FS}  &  0-15  &  16-20 &  \textbf{All}  & \textbf{FS}  &  0-5   &  6-20  &  \textbf{All}  & \textbf{FS} &  0-10  &  11-20 &  \textbf{All}  & \textbf{FS} \\ \boldhline
		FT          & 05.89  & 40.47  & 14.12  & \forgetscore{67.57}  
					  & 04.56  & 01.52  & 03.84 & \forgetscore{77.85}  
					  & 12.05  & 05.66  & 07.49 & \forgetscore{74.20}
					  & 06.45  & 01.12  & 03.91  & \forgetscore{77.78} \\ \hline
	 
	 CLoRA (FT)    & 04.69  & 43.47  & 13.94 & \forgetscore{68.72}  
	 						 & 04.44  & 02.11  & 03.88 & \forgetscore{78.78}  
	 						 & 11.78  & 07.00  & 08.37 & \forgetscore{74.29} 
	 						 & 06.38  & 01.21  & 03.92  & \forgetscore{78.74} \\ \boldhline
		JT          & 82.53  & 79.02  & 81.69 & 0.00 & 82.53  & 79.02  & 81.69 & 0.00 & 80.27  & 82.26  & 81.69 & 0.00 & 81.57  & 81.83  & 81.69 & 0.00 \\ \hline
		CLoRA (JT)     & 83.52  & 79.89  & 82.66 & 0.00 & 83.52  & 79.89  & 82.66 & 0.00 & 81.13  & 83.27  & 82.66 & 0.00 & 83.02  & 82.26  & 82.26 & 0.00 \\ \boldhline
		MiB        & 77.52  & 49.73  & \textbf{70.91} & \forgetscore{10.78} 
					  & 75.22  & 19.47  & 61.95 & \forgetscore{19.74}  
					  & 63.23  & 42.18  & 48.19 & \forgetscore{33.50} 
					  & 26.47  & 19.70  & 23.24  & \forgetscore{58.45} \\ \hline
		
		MiB (TL) & 21.03  & 04.62  & 17.12 & \forgetscore{64.57}  
						& 17.76  & 03.15  & 14.28 & \forgetscore{67.41} 
						& 15.21  & 04.32  & 07.43 & \forgetscore{74.26} 
						& 00.00  & 03.43  & 01.63  & \forgetscore{80.06} \\ \hline
		
		CLoRA Reinit  & 71.14 & 52.58  & 66.34 & \forgetscore{16.32} 
								& 80.82  & 31.47  & 69.07 & \forgetscore{13.59} 
								& 62.68  & 45.06  & 50.09 & \forgetscore{32.56} 
								& 22.66  & 23.17  & 22.90 & \forgetscore{59.76} \\ \hline

		CLoRA    & 74.17  & 56.57  & 70.39 & \forgetscore{12.27} 
						& 81.29  & 34.41  & \textbf{70.13} & \forgetscore{12.53} 
						& 69.92  & 45.50  & \textbf{52.47} & \forgetscore{30.19} 
						& 31.38  & 29.22  & \textbf{30.35}  & \forgetscore{52.31} \\ \boldhline
	\end{tabular}
\end{adjustbox}
\label{tab:voc}	
\end{table*}

\subsection{Baselines and Implementation}
\label{sec:implmentation}
To evaluate the efficacy of CLoRA, a PECL method, we compare it against full fine-tuning. 
Full fine-tuning refers to the standard training where the entire model with all the parameters are retrained for each task. 
These approaches include the continual learning baselines of fine-tuning and joint training. 
Fine-tuning (FT) involves incrementally learning new tasks with the trained model, without any explicit intervention to mitigate catastrophic forgetting.
Joint training (JT) or offline training uses all the data to train the model in a single step.
Since there is no incremental learning, it circumvents forgetting and forms the upper bound for comparison. 
Primarily , we compare against Modeling the Background (MiB) \citep{MiB}, which uses full fine-tuning and highlight the efficacy of CLoRA, utilizing the same knowledge distillation loss.
We further include comparison with SATS \citep{SATS}, SSUL \citep{SSUL} and RCIL \cite{RCIL} to demonstrate the versatility of CLoRA.
\newline\newline
The segmentation network consists of a Vision Transformer (ViT) \citep{ViT} as the encoder, and we use the corresponding LoRA implementation by \cite{loraVIT}. 
The decoder and the classifier consist of a single convolutional layer, and we use the CL framework by \cite{MiB}.
For training the full fine-tuning models we use the default hyperparameters defined by \cite{MiB}.
For training with CLoRA, we use a batch size of 6 with a higher learning rate of 0.04 for the initial task, and for subsequent tasks we use a learning rate of 0.001 for smaller increments of single classes and 0.005 for all other increments.
We use a rank $r=32$ for LoRA, which amounts to 1.04\% of the total trainable parameters of the model. LoRA is applied only to the encoder, while the decoder and the classifiers are fine-tuned.
We study the effect of rank $r$ on the performance of the model and determine $r=32$ is sufficient for almost all experiments.
All models are trained for 30 epochs on each task and are evaluated using mean IoU. 
We present results on both the initial and incremental tasks to analyze the approach's balance between learning and retaining information.
\subsection{Task Settings and Evaluation}
\label{sec:eval}
We present results from various CL tasks across three datasets. 
In class-incremental learning, the tasks follow the format $init$-$inc$, where $init$ is number of classes learned initially and the $inc$ is number of classes learned in each increment. 
The steps are repeated until all classes are learned.

\subsubsection{PASCAL VOC} 
We present four CL experiments using the 21 classes in PASCAL, with different sequence lengths: \textit{15-5} (2 steps), \textit{15-1} (6 steps), \textit{5-3} (6 steps), and \textit{10-1} (11 steps).
The results from these experiments are presented in \cref{tab:voc}, and we observe that for most tasks except \textit{15-5}, CLoRA surpasses MiB. 
CLoRA demonstrates greater effectiveness in longer and more challenging sequences of tasks, as observed in \textit{15-1}, \textit{5-3}, and \textit{10-1}.
Notably, CLoRA is more adept in learning new tasks across all experiments, despite MiB achieving slightly better overall results in the \textit{15-5} setting.
In the remaining experiments, CLoRA significantly outperforms MiB in retaining previous knowledge. 

\begin{table*}
	\caption{Results on ADE20K \citep{ADE} dataset after learning all tasks.}
	\begin{adjustbox}{width=\textwidth}
		\centering
		\begin{tabular}{c||c|c|c|c||c|c|c|c||c|c|c|c||c|c|c|c}
			\boldhline
			\multirow{2}{*}{\textbf{Method}} & \multicolumn{4}{c||}{\textbf{100-50}} & \multicolumn{4}{c||}{\textbf{50-50}} & \multicolumn{4}{c||}{\textbf{25-25}} & \multicolumn{4}{c}{\textbf{100-10}} \\ \cline{2-17}
			& 0-100  & 101-150 & \textbf{All} & \textbf{FS} & 0-50 & 51-150 & \textbf{All} & \textbf{FS} & 0-25 & 26-150 & \textbf{All} & \textbf{FS} & 0-100 & 101-150 & \textbf{All} & \textbf{FS} \\ \boldhline
			
			FT & 00.09 & 10.55 & 03.58 & \forgetscore{40.68} 
			& 00.01 & 06.47 & 04.31 & \forgetscore{39.95} 
			& 00.00 & 03.47 & 02.89 & \forgetscore{41.37} 
			& 00.00 & 00.20 & 00.06 & \forgetscore{44.20} \\ \hline
			
			CLoRA (FT) & 00.00 & 21.69 & 07.23 & \forgetscore{34.12} 
			& 00.00 & 09.60 & 06.40 & \forgetscore{34.95} 
			& 00.00 & 04.94 & 04.12 & \forgetscore{37.23} 
			& 00.00 & 01.75 & 00.58 & \forgetscore{40.77} \\ \boldhline
			
			JT & 49.53 & 33.73 & 44.26 & 0.00 
			& 57.38 & 37.71 & 44.26 & 0.00 
			& 67.40 & 39.64 & 44.26 & 0.00 
			& 49.53 & 33.73 & 44.26 & 0.00 \\ \hline
			
			CLoRA (JT) & 48.85 & 26.35 & 41.35 & 0.00 
			& 56.83 & 33.61 & 41.35 & 0.00 
			& 66.35 & 36.35 & 41.35 & 0.00 
			& 48.85 & 26.35 & 41.35 & 0.00 \\ \boldhline		
			
			MiB & 46.63 & 20.80 & 38.02 & \forgetscore{06.24} 
			& 50.12 & 21.27 & 30.89 & \forgetscore{13.37} 
			& 62.33 & 27.25 & 33.09 & \forgetscore{11.17} 
			& 42.17 & 08.22 & \textbf{30.85} & \forgetscore{13.41} \\ \hline
			
			CLoRA & 44.43 & 25.52 & \textbf{38.13} & \forgetscore{03.22} 
			& 49.05 & 25.85 & \textbf{33.58} & \forgetscore{07.77} 
			& 61.02 & 30.06 & \textbf{35.22} & \forgetscore{06.13} 
			& 39.27 & 13.47 & 30.67 & \forgetscore{10.68} \\ \boldhline
		\end{tabular}
	\end{adjustbox}
	\label{tab:ade}
\end{table*}

\begin{table*}[t]
	\caption{Results on Cityscapes \citep{cityscapes} dataset after learning all tasks.}
	\begin{adjustbox}{width=\textwidth}
		\begin{tabular}{c||c|c|c|c||c|c|c|c||c|c|c|c||c|c|c|c}
			\boldhline
			\multirow{2}{*}{\textbf{Method}} & \multicolumn{4}{c||}{\textbf{14-5}} & \multicolumn{4}{c||}{\textbf{14-1}} & \multicolumn{4}{c||}{\textbf{7-3}} & \multicolumn{4}{c}{\textbf{10-1}} \\ \cline{2-17}
			& 1-14 & 15-19 & \textbf{All} & \textbf{FS} & 1-14 & 15-19 & \textbf{All} & \textbf{FS} & 1-7 & 8-19 & \textbf{All} & \textbf{FS} & 1-10 & 11-19 & \textbf{All} & \textbf{FS} \\ \boldhline
			
			FT & 00.00 & 00.13 & 00.03 & \forgetscore{59.35} 
			& 00.00 & 00.00 & 00.00 & \forgetscore{59.38} 
			& 00.00 & 02.25 & 01.42 & \forgetscore{57.96} 
			& 00.00 & 03.18 & 01.51 & \forgetscore{57.87} \\ \hline
			
			CLoRA (FT) & 00.00 & 24.78 & 06.52 & \forgetscore{54.25} 
			& 00.00 & 03.31 & 00.87 & \forgetscore{59.90} 
			& 00.00 & 07.54 & 04.76 & \forgetscore{56.01} 
			& 00.00 & 00.01 & 00.00 & \forgetscore{60.77} \\ \boldhline
			
			JT & 61.83 & 52.52 & 59.38 & 0.00 
			& 61.83 & 52.52 & 59.38 & 0.00 
			& 58.07 & 60.14 & 59.38 & 0.00 
			& 59.35 & 59.71 & 59.38 & 0.00 \\ \hline
			
			CLoRA (JT) & 61.41 & 58.99 & 60.77 & 0.00 
			& 60.41 & 58.99 & 60.77 & 0.00 
			& 56.34 & 63.35 & 60.77 & 0.00 
			& 57.87 & 63.98 & 60.77 & 0.00 \\ \boldhline
			
			MiB & 59.73 & 08.93 & 46.36 & \forgetscore{13.02} 
			& 60.20 & 07.47 & 46.32 & \forgetscore{13.06} 
			& 49.26 & 27.61 & 35.58 & \forgetscore{23.80} 
			& 55.44 & 30.41 & \textbf{43.59} & \forgetscore{15.79} \\ \hline
			
			CLoRA & 60.78 & 36.14 & \textbf{54.30} & \forgetscore{06.47} 
			& 61.57 & 13.01 & \textbf{48.79} & \forgetscore{11.98} 
			& 55.89 & 43.73 & \textbf{48.21} & \forgetscore{12.56} 
			& 56.36 & 23.86 & 40.96 & \forgetscore{19.81} \\ \boldhline
		\end{tabular}
	\end{adjustbox}
	\label{tab:cs}
\end{table*}

\begin{table}[t]
	\caption{Results on PASCAL VOC \citep{PASCAL} dataset after learning all tasks using SegFormer \citep{segformer}. $\ast$ indicates rank $r=20$. }
	\centering
	\begin{tabular}{c||c|c|c|c}
		\boldhline
		\textbf{Method} & \textbf{15-5}  & \textbf{15-1} &  \textbf{5-3} & \textbf{10-1} \\\boldhline
		
		MiB \citep{MiB}  	& \textbf{69.90}  & 58.82  & \textbf{52.05} & 40.69 \\ \hline
		MiB + CLoRA  		& 69.83  & \textbf{59.89}  & 51.48 &  \textbf{43.32} \\ \boldhline
		SATS \citep{SATS} 	& 69.23  & \textbf{61.49}  & \textbf{55.00} & 39.67 \\ \hline
		SATS + CLoRA 		& \textbf{69.65}  & 60.83 & 52.29* & \textbf{40.23}* \\ \boldhline
	\end{tabular}
	\label{tab:pascal_segformer}
\end{table}

\subsubsection{ADE20K}

Leveraging the large number of classes in the ADE, we design four experiments where each step introduces a significant number of new classes. These include \textit{100-50} (2 steps), \textit{50-50} (3 steps), \textit{25-25} (6 steps), \textit{100-10} (6 steps).
The results are presented in \cref{tab:ade}, in both the \textit{100-50} and \textit{100-10} settings, the results achieved by MiB and CLoRA are relatively similar, with CLoRA achieving slightly better performance in the \textit{100-50} setting and MiB in the \textit{100-10} setting.
In the remaining \textit{50-50} and \textit{25-25} settings, CLoRA outperforms MiB by a significant margin.
Once again, the effectiveness of CLoRA in learning new classes is evident across all experiments. 
However, with joint training, CLoRA underperforms compared to full fine-tuning, possibly due to the dataset's large size. This is further illustrated by examining the effect of rank on learning under \cref{sec:addn_exp}. 
With fine-tuning, we can observe complete overwriting of information from the initial task, both with CLoRA and default fine-tuning. However, we can observe CLoRA being able to learn new classes to a greater extent.

\subsubsection{Cityscapes}
Utilizing the Cityscapes dataset, we replicate experiments similar to those conducted with PASCAL, resulting in the following experiments: \textit{14-5} (2 steps), \textit{14-1} (5 steps), \textit{7-3} (5 steps), and \textit{10-1} (10 steps).
Unlike PASCAL, where images typically feature atmost few classes, Cityscapes contains multiple recurring classes such as road, sky, buildings, and vehicles, making it a much more challenging dataset.
From the results presented in \cref{tab:cs}, we observe that almost across all experiments, barring \textit{10-1}, CLoRA surpasses MiB.
Notably, even with joint training, CLoRA performs better compared to full fine-tuning.
The results from the fine-tuning approach exhibit the lowest performance in Cityscapes compared to the other two datasets. Besides completely overwriting previous task knowledge, it fails to learn new classes. This can be attributed to the fact that the classes learned incrementally are underrepresented, and the model lacks sufficient data to learn effectively.

\subsection{Additional Experiments}
\label{sec:addn_exp}

\subsubsection{Extending CLoRA} \label{sec:experiments:additional:extension}
We demonstrate the model- and approach-agnostic nature of CLoRA, which allows for the use of any distillation method to transfer knowledge between tasks. 
In this experiment, we employ the SegFormer \citep{segformer} network and MiT-B1 with MiB \citep{MiB} and SATS \citep{SATS} to illustrate CLoRA’s adaptability. 
We use rank $r=16$ which corresponds to 4.91\% trainable parameters. 
While CLoRA exhibits maximum benefit with larger networks, this experiment highlights its suitability even for smaller networks. 
All previously discussed advantages of CLoRA regarding efficiency are preserved here, albeit proportionally. 
We repeat all tasks from the PASCAL VOC dataset \citep{PASCAL} and the results are presented in \cref{tab:pascal_segformer}.

\subsubsection{CLoRA vs Frozen Encoder} \label{sec:experiments:additional:frozen}

We highlight the efficacy of CLoRA, by comparing it with a frozen encoder, and only fine-tuning the decoder for CIL, similar to transfer-learning based CL.
Typically, encoders use pretrained networks on ImageNet \citep{ImageNet} or from SAM \citep{SAM} in our case, which holds the capability to generalize to other datasets.
We repeat the PASCAL experiments with a frozen encoder. 
The results are presented in \cref{tab:voc} as MiB (TL).
The model fails to learn adequately even for the initial task, presumably due to the domain gap between the pretrained data and the task data, resulting in poor performance on subsequent tasks. 
By fine-tuning only 1\% of parameters, CLoRA outperforms significantly, surpassing even full fine-tuning which uses 100\% of the parameters.

\begin{table}[t]
	\centering
	\caption{Performance across varying ranks for joint training. Cityscapes shows a linear performance increase with rank, PASCAL VOC results fluctuate, and ADE displays CLoRA underperforming compared to full fine-tuning.}
	\label{tab:ranks}
	\renewcommand{\arraystretch}{1.2}
	\setlength{\tabcolsep}{10pt}
	\begin{tabular}{l||c|c|c|c|c||c}
		\boldhline
		\multirow{2}{*}{\textbf{Dataset}} & \multicolumn{5}{c||}{\textbf{Ranks}} & \multirow{2}{*}{\textbf{Baseline}} \\
		\cline{2-6}
		& \textbf{16} & \textbf{32} & \textbf{64} & \textbf{96} & \textbf{128} & \\
		\boldhline
		PASCAL VOC \citep{PASCAL}      & 81.50 & 82.66 & 81.53 & 82.85 & 82.03 & 81.69 \\ \hline
		Cityscapes \citep{cityscapes}  & 59.79 & 60.77 & 60.87 & 61.56 & 61.23 & 59.38 \\ \hline
		ADE20K  \citep{ADE}    & 40.98 & 41.35 & 42.53 & 41.91 & 41.64 & 44.26 \\
		\boldhline
	\end{tabular}
\end{table}

\begin{figure}[t]
	\centering
	\includegraphics[width=0.97\textwidth]{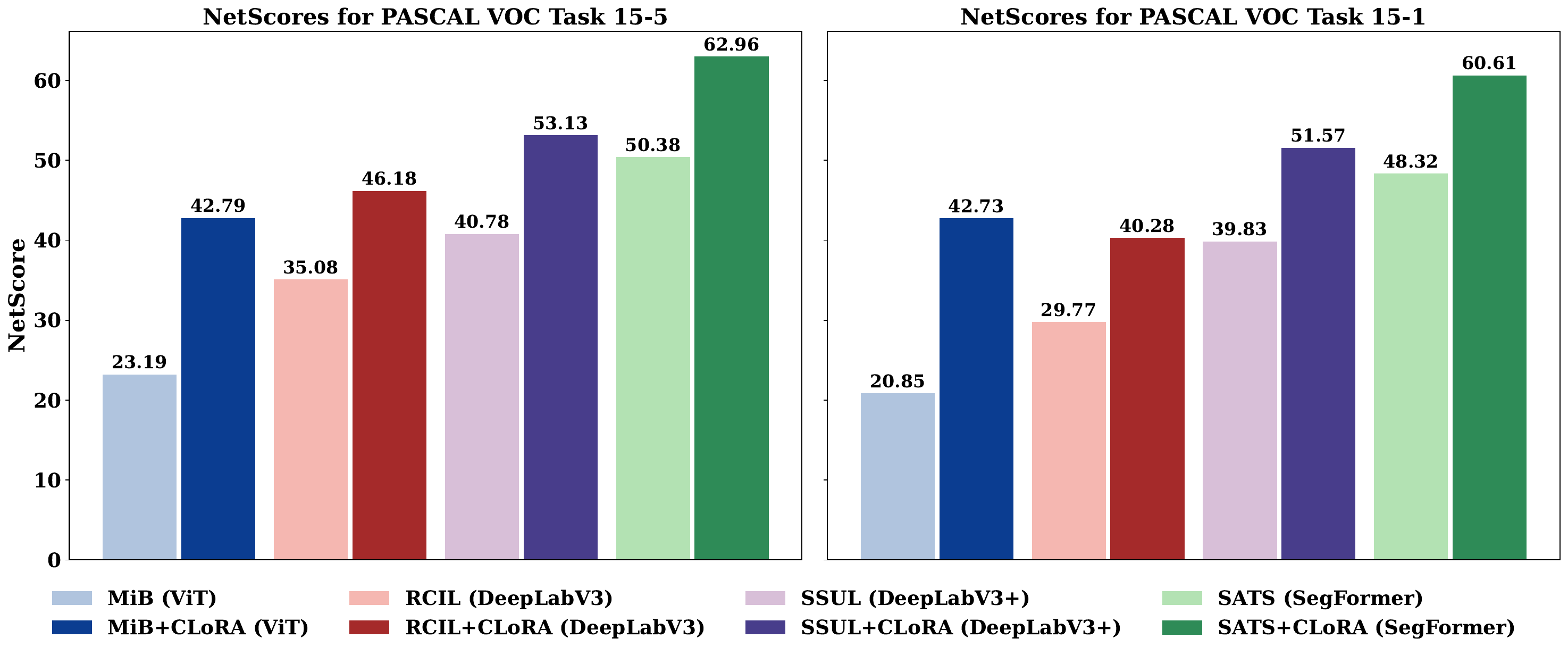}
	\caption{NetScore results across CL scenarios on PASCAL VOC \citep{PASCAL} using different baselines and networks. CLoRA improves NetScore substantially across network sizes, enhancing resource efficiency1-10.}
	\label{fig:netscore}
\end{figure}

\subsubsection{Reinitializing LoRA} \label{sec:experiments:additional:reinit}
For the initial task, the LoRA parameters $A$ and $B$ for all query and value matrices in the encoder are initialized to the default values and then updated through training.
These updated LoRA weights are used when learning the subsequent tasks, while preserving the original pretrained weights $W$.
In this experiment, we investigate the effects of reinitializing the LoRA weights after each task. 
This involves merging LoRA with the pretrained weights after learning each task $t$, resulting in updated weights $W_t$. 
For the subsequent task, these updated weights $W_t$ are used instead of the original weights $W$, and the LoRA weights are recreated with default values.
In the seventh row of \cref{tab:voc}, we observe that this approach does not significantly influence the performance, and even performs sub-optimally in certain cases. 

\subsubsection{Effect of LoRA Rank} \label{sec:experiments:additional:ranks}
The rank $r$ serves as a hyperparameter that determines the number of trainable parameters. 
Across nearly all experiments, we observe that a rank of $r=32$, which represents approximately 1\% of the total parameters, is adequate for learning all tasks. 
The performance across varying ranks and for offline joint training is depicted in \cref{tab:ranks} across the three datasets.
More results for incremental settings are in the appendix.
We observe that for Cityscapes, the results increase linearly with an increase in rank. 
For PASCAL, the results seem to be saturated, with performance fluctuating across different ranks. 
Notably, in the case of offline training with ADE, CLoRA underperforms compared to full fine-tuning. 
This discrepancy could be attributed to the larger size of the dataset relative to the other two datasets.

\subsection{Efficiency aspects of CLoRA} \label{sec:clora_fft}

Currently, CL methods focus on task performance and the ability to retain knowledge across tasks without forgetting.
However, there is a growing emphasis on the efficiency aspects of CL methods beyond forgetting for practical applicability of CL in resource constrained environments \citep{clEfficiency, onlineCL}. 
NetScore \citep{NetScore} provides a comprehensive metric combining performance $a_N$, network size $p_N$, and the computational complexity $m_N$.
Following previous studies \citep{clEfficiency, onlineCL, llStreamLearning}, we also use NetScore for a holistic and effective evaluation of CL models, emphasizing the importance of resource efficiency alongside performance. The modified NetScore $\Omega_N$ for CL training is calculated as:
\begin{equation}
	\Omega_N = 20 \log \left( \frac{a_N^{\alpha}}{p_N^{\beta} \cdot m_N^{\gamma}} \right)
\end{equation}
where $a_N$ is the final mIoU after learning all tasks, $p_N$ is the number of parameters in millions, and $m_N$ is the number of multiply–accumulate (MAC) operations.
According to \citep{NetScore}, $m_N$ is measured during inference; however, we consider it in the training phase, since the focus of CL is updating a model.
We use the default values of $\alpha = 2, \beta = \gamma = 0.5$.
We evaluate the impact of CLoRA using NetScore across multiple continual learning methods (MiB \citep{MiB}, SATS \citep{SATS}, SSUL \citep{SSUL} and RCIL \citep{RCIL}) by comparing each baseline with the corresponding CLoRA-augmented method.
We demonstrate the effectiveness of CLoRA across a wide range of networks including Vision Transformer \citep{ViT}, SegFormer \citep{segformer}, DeepLabV3 \citep{DeepLabV3} and DeepLabV3+ \citep{DeepLabV3Plus}.
While certain continual learning methods such as SSUL \citep{SSUL} and RCIL \citep{RCIL} are inherently efficient by updating a subset of parameters during the incremental steps, CLoRA improves the efficiency by further reducing the number of trainable parameters.
To compute NetScore, the number of parameters is averaged between the full set used in the initial step and the reduced set used during incremental learning, providing a fair measure of resource efficiency.

\cref{fig:netscore} presents the NetScore results across different continual learning baselines and networks for two tasks from PASCAL VOC \citep{PASCAL}. 
For the larger ViT-based network, we observe a substantial increase in the NetScore with CLoRA, highlighting its effectiveness in optimizing larger models. 
For the smaller networks, which already attains a relatively high NetScore, CLoRA further provides an improvement.
CLoRA consistently demonstrates advantages across network sizes, enhancing resource efficiency without compromising performance relative to fully trained models.
We analyse the trade-off between performance and efficiency by plotting the Pareto front of mIoU \vs trainable parameters for 15-5 and 15-1 task settings on PASCAL VOC \citep{PASCAL}. 

\begin{figure}[tb]
	\centering
	\includegraphics[width=0.98\textwidth]{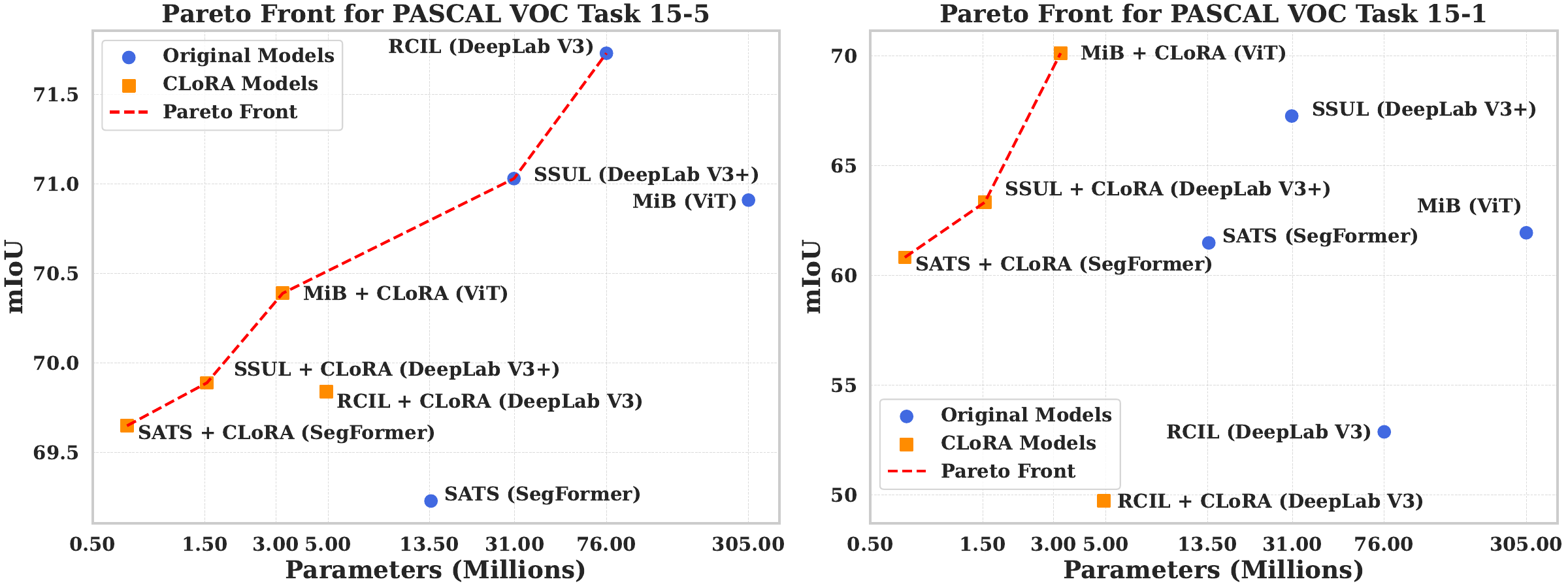}
	\caption{Pareto Front of mIoU \vs Trainable Parameters on PASCAL VOC \citep{PASCAL}. The plots compare the performance-efficiency trade-off for models and the corresponding CLoRA augmented methods.}
	\label{fig:pareto_front}
\end{figure}

The Pareto front in \cref{fig:pareto_front} highlights not just the best-performing methods, but those that offer the most favourable balance between accuracy and efficiency.
While RCIL achieves the highest mIoU and appears Pareto-optimal in the 15-5 setting, its CLoRA counterpart is dominated by more efficient alternatives such as SSUL+CLoRA and MiB+CLoRA, rather than by RCIL itself.
Notably, across all methods, the CLoRA-augmented variants consistently improve the efficiency–performance trade-off and are never dominated by their corresponding baselines.
In the 15-1 setting, both RCIL and RCIL+CLoRA are dominated, suggesting that the underlying method is the limiting factor rather than the use of CLoRA. 
This reinforces that while CLoRA enhances parameter efficiency, the overall performance is still bounded by the effectiveness of the baseline method.
\newline\newline

\section{Conclusion}
\label{sec:conclusion}
In this work, we present Continual Learning with Low-Rank Adaptation (CLoRA), a parameter-efficient continual learning (PECL) method. 
CLoRA utilizes LoRA for incrementally learning new tasks, using a small fraction of parameters instead of training all the parameters.
In contrast to existing PECL methods focusing on image classification with task-specific modules, we discuss the constraints in extending it to class-incremental segmentation.
Addressing these challenges, we design CLoRA as a single module reused across all tasks and updated using knowledge distillation. 
We demonstrate the effectiveness of CLoRA, achieving results on-par, and surpassing the baselines where all parameters are updated.
This introduces a novel and efficient training process for continual learning with limited resources, without sacrificing model performance.
One potential limitation of CLoRA could be in handling larger datasets, although this is less of a concern in continual learning, where increments are typically small.

\section*{Acknowledgments}
This work was partially funded by the Federal Ministry of Education and Research Germany under the projects DECODE (01IW21001) and COPPER (01IW24009).

\bibliography{main}
\bibliographystyle{collas2025_conference}

\clearpage
\appendix
\section*{Appendix}

We first detail the issues with individual, task-specific modules, \ie merging of predictions.
We assess the robustness of a single LoRA module to handle both input and output distribution shifts.
Then, we provide additional visual results for our experiments.
Lastly, we perform additional experiments with varied LoRA ranks.

\section{Merging Task-specific Modules}
\begin{figure}[b]
	\centering
	\begin{adjustbox}{width=\textwidth}
		\setlength{\tabcolsep}{1pt}
		\renewcommand\arraystretch{0.1}
		\begin{tabular}{cccccc}
			\small
			\textbf{Image} & \textbf{GT} & \textbf{Task 0} & \textbf{Task 2}   & \textbf{Task 3} & \textbf{Task 7} \\
			
			\includegraphics[width=.16\linewidth]{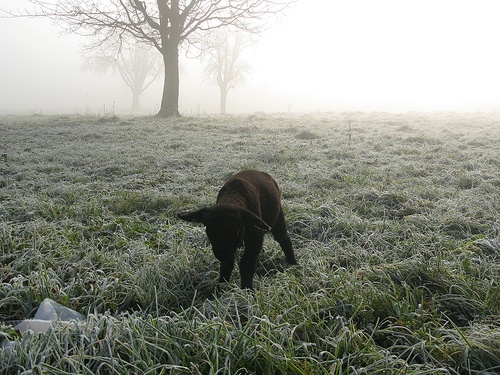} &
			\includegraphics[width=.16\linewidth]{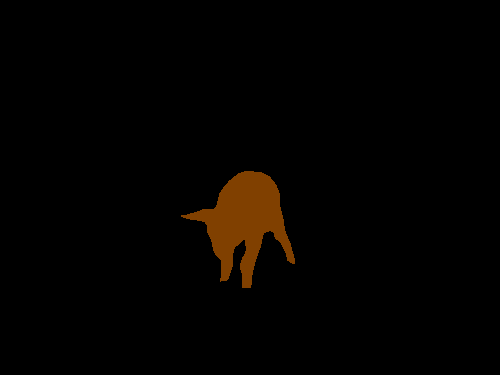} &
			\includegraphics[width=.16\linewidth]{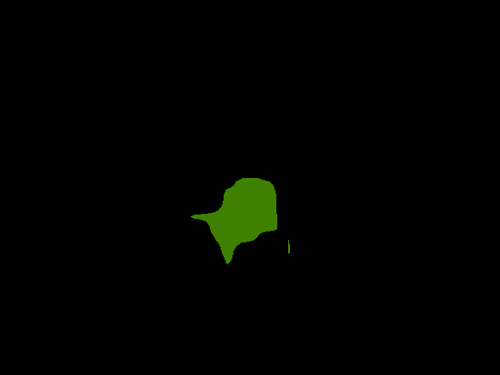} &
			\includegraphics[width=.16\linewidth]{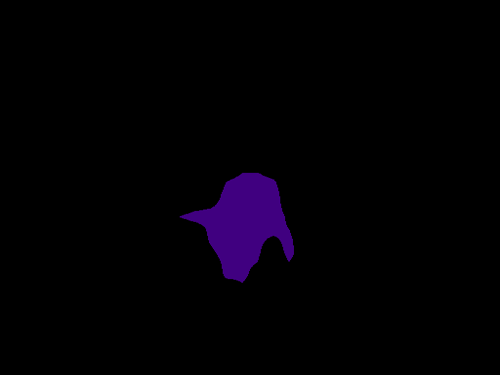} &
			\includegraphics[width=.16\linewidth]{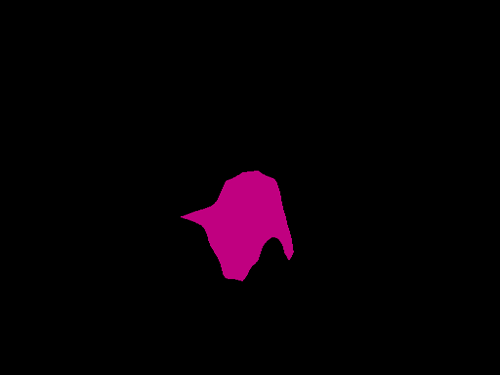} &
			\includegraphics[width=.16\linewidth]{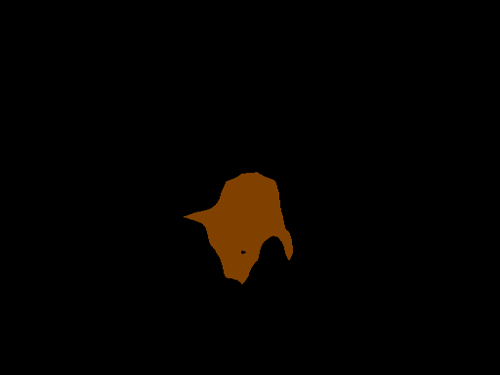} \\ [0.5ex]
			
			\multicolumn{6}{c}{
				\includegraphics[width=0.75\textwidth]{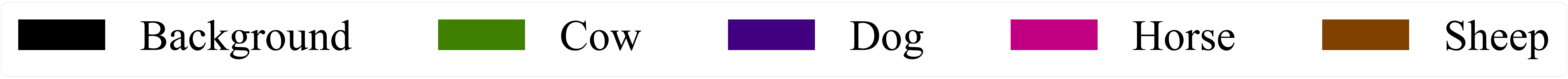}
			} \\
			
		\end{tabular}
	\end{adjustbox}
	
	\caption{Conflicting predictions from task-specific modules on the PASCAL VOC \citep{PASCAL} dataset using the \textit{10-1} setting, in which multiple modules have conflicting predictions.}
	\label{fig:supp:merging_preds}
\end{figure}

Individual adapters per task are only aware of the classes they have learned.
Naturally it might occur that more than one module predicts multiple non-background classes for the same pixel.
In such cases, which occur especially when semantically and visually similar classes are learned in different tasks, the individual predictions need to be merged before a final class is derived.
\Cref{fig:supp:merging_preds} illustrates this challenge using PASCAL VOC \citep{PASCAL} tasks. In the \textit{10-1} setting, for the class \textit{sheep} learned in task 7, the previous modules that were trained on other animal classes predict the specific animal associated with their respective tasks.
This error arises because the animal classes are visually similar, and the modules trained on earlier tasks have not encountered images of \textit{sheep} to discriminate between \textit{sheep} and the animal class learned in their respective tasks.
Resolving the conflict is non-trivial.
A naive, inferior solution could be achieved by concatenation of all logits.
\Ie during inference, the unnormalized logits of all task-specific classifiers are concatenated and soft-maxed. Afterwards, the most probable class is selected.
However, this has limitations, \eg the task-wise class distributions are not calibrated (different numbers of classes per task).

\section{Robustness to Distribution Shifts}

This work focuses on class-incremental learning which involves sequentially learning new classes. 
This results in a shift in the output distribution, while the input distribution remains constant.
However, in real-world scenarios, domain shift may occur due to varying weather, lighting, or geographical locations.
Such a novel incremental setting with both semantic shift (new classes) and domain shift (new domains) has been previously studied by \cite{LwS, PSS}.
While this is not the primary focus of our work, we recognize the importance of assessing whether our approach with CLoRA, can remain effective under such shifts.
To this end, we perform an experiment using Cityscapes \citep{cityscapes} as the base task (initial step), and ACDC \citep{ACDC} which consists of adverse domains for the incremental tasks.
Both datasets share the same set of semantic classes, which allows us to design a class-incremental learning setup while the differing visual conditions introduce an input domain shift.
We use MiB \citep{MiB} as the baseline continual learning method and report results on ACDC with and without CLoRA in \cref{tab:hybrid_il}. 
Notably, we use a single LoRA module across all tasks, without any task-specific adaptation for different domains.
Despite the additional domain gap, CLoRA exhibits the same performance trends relative to MiB as observed in the class-incremental experiments using only Cityscapes \citep{cityscapes}.

\begin{table*}[t]
	\caption{Results on class-incremental learning with varying input domains using Cityscapes \citep{cityscapes} and ACDC \citep{ACDC}.}
	\begin{adjustbox}{width=\textwidth}
		\begin{tabular}{c||c|c|c|c||c|c|c|c||c|c|c|c}
			\boldhline
			\multirow{2}{*}{\textbf{Method}} & \multicolumn{3}{c|}{\textbf{14-5}} & \multicolumn{3}{c|}{\textbf{14-1}} & \multicolumn{3}{c|}{\textbf{7-3}} & \multicolumn{3}{c}{\textbf{10-1}} \\ \cline{2-13}
			& 1-14 & 15-19 & \textbf{All}  
			& 1-14 & 15-19 & \textbf{All} 
			& 1-7 & 8-19 & \textbf{All} 
			& 1-10 & 11-19 & \textbf{All} \\ \boldhline 
			
			MiB & 48.11 & 02.08 & 36.00 
			& 48.31 & 04.05 & 36.66 
			& 42.15 & 19.89 & 28.09 
			& 44.09 & 18.46 & \textbf{31.95} \\ \hline
			
			CLoRA & 47.96 & 16.89 & \textbf{39.79} 
			& 49.06 & 03.75 & \textbf{37.14} 
			& 47.16 & 26.05 & \textbf{33.83} 
			& 43.71 & 16.67 & 30.90 \\ \boldhline
		\end{tabular}
	\end{adjustbox}
	\label{tab:hybrid_il}
\end{table*}

\section{Additional Visualizations}
\begin{figure*}[t]
	\begin{adjustbox}{width=\textwidth}
		\centering
		\setlength{\tabcolsep}{1pt}
		\begin{tabular}{ccccccc}
			\small
			\renewcommand\arraystretch{0.1}
			\textbf{Image} & \textbf{GT} & \textbf{JT} & \textbf{CLoRA (JT)} & \textbf{MiB} & \textbf{CLoRA} & \\
			
			\includegraphics[width=.2\linewidth]{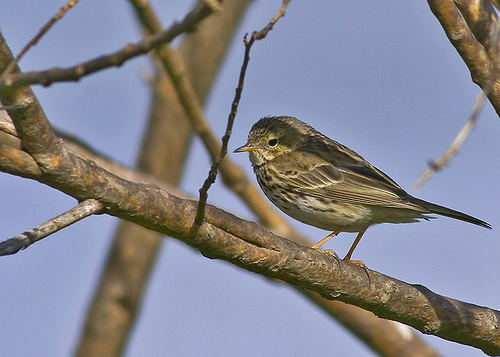} &
			\includegraphics[width=.2\linewidth]{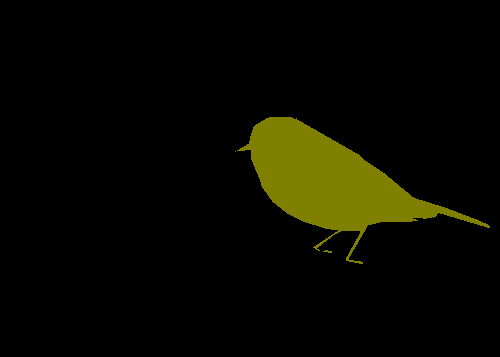} &
			\includegraphics[width=.2\linewidth]{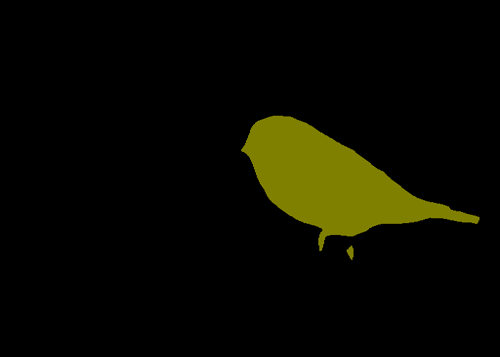} &
			\includegraphics[width=.2\linewidth]{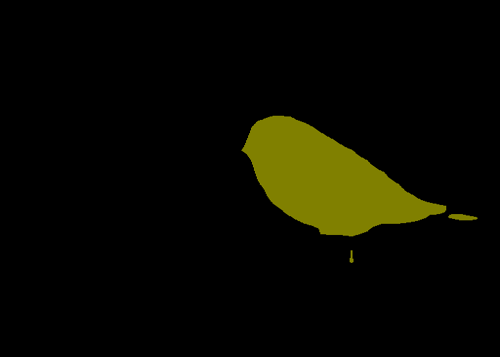} &
			\includegraphics[width=.2\linewidth]{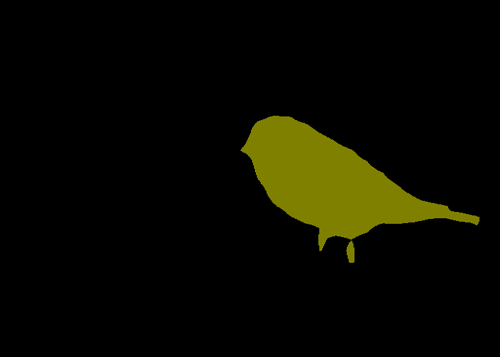} &
			\includegraphics[width=.2\linewidth]{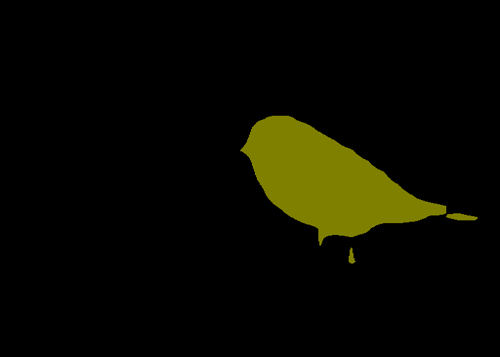} & \raisebox{1.25cm}{\rotatebox{-90}{\textbf{15-5}}} \\[-0.5ex]
			
			\includegraphics[width=.2\linewidth]{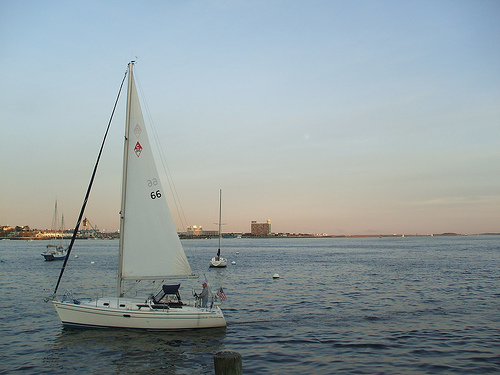} &
			\includegraphics[width=.2\linewidth]{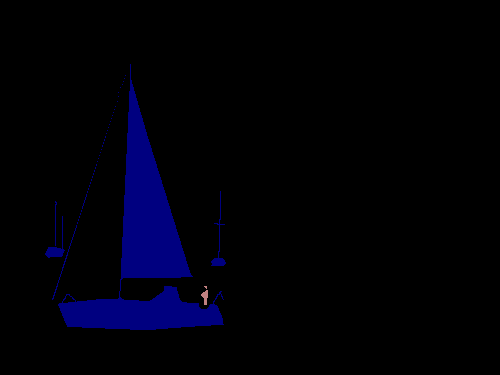} &
			\includegraphics[width=.2\linewidth]{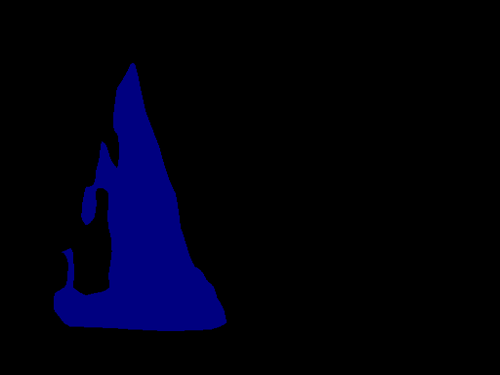} &
			\includegraphics[width=.2\linewidth]{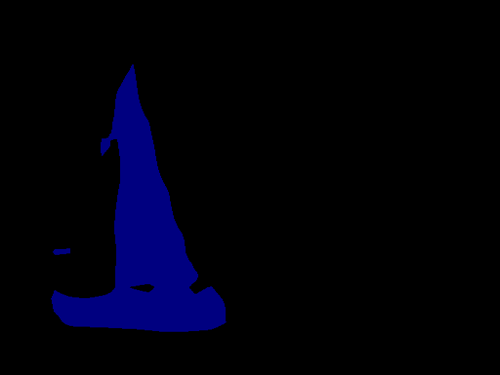} &
			\includegraphics[width=.2\linewidth]{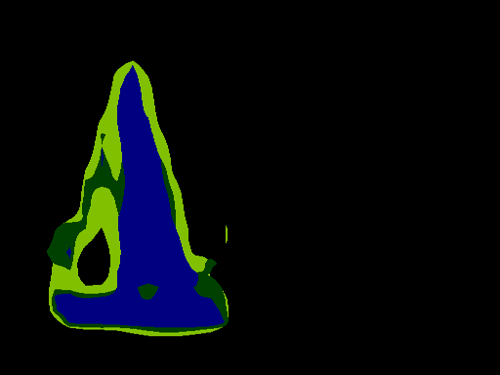} &
			\includegraphics[width=.2\linewidth]{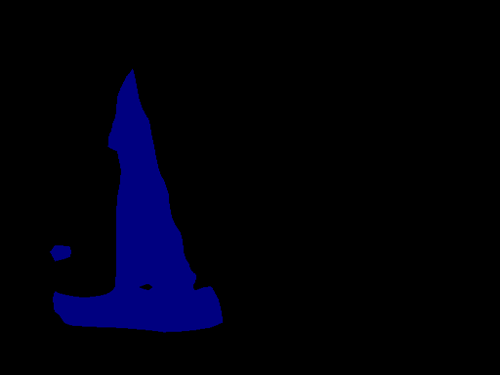} & \raisebox{1.25cm}{\rotatebox{-90}{\textbf{15-1}}} \\[-0.5ex]
			
			\includegraphics[width=.2\linewidth]{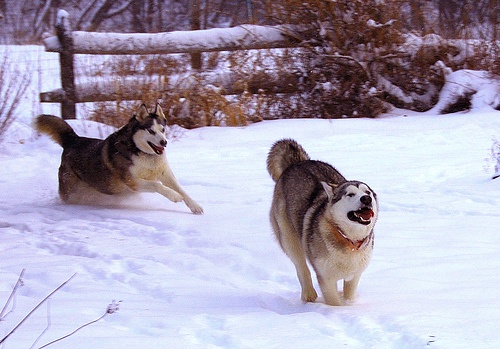} &
			\includegraphics[width=.2\linewidth]{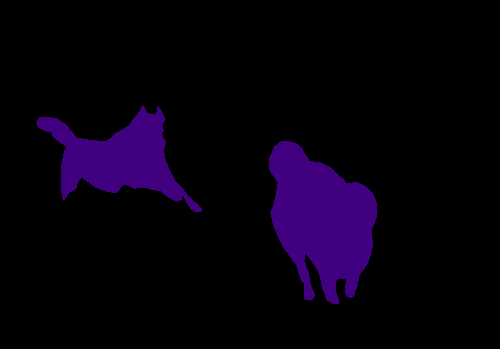} &
			\includegraphics[width=.2\linewidth]{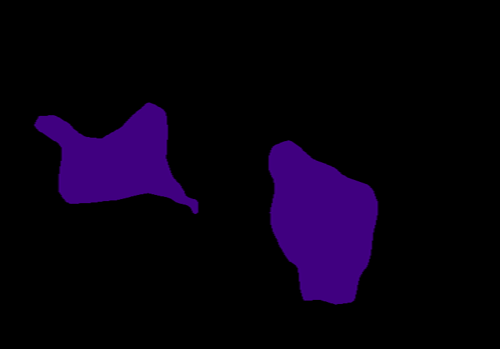} &
			\includegraphics[width=.2\linewidth]{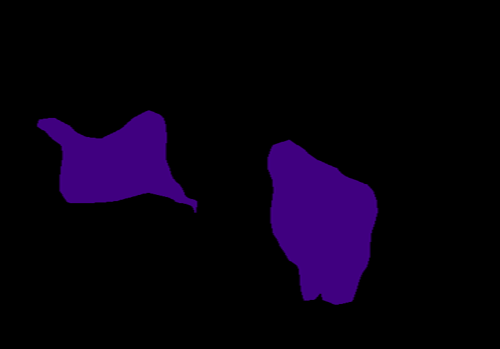} &
			\includegraphics[width=.2\linewidth]{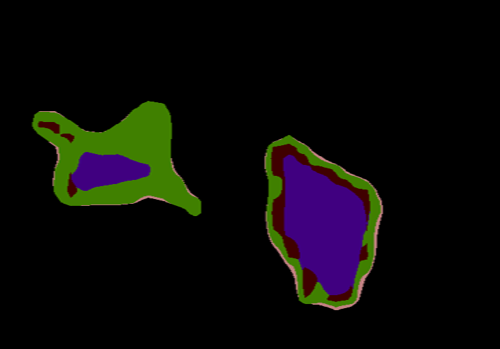} &
			\includegraphics[width=.2\linewidth]{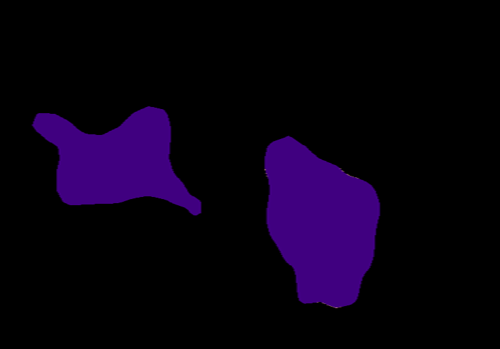} & \raisebox{1.4cm}{\rotatebox{-90}{\textbf{5-3}}} \\[-0.5ex]
			
			\includegraphics[width=.2\linewidth]{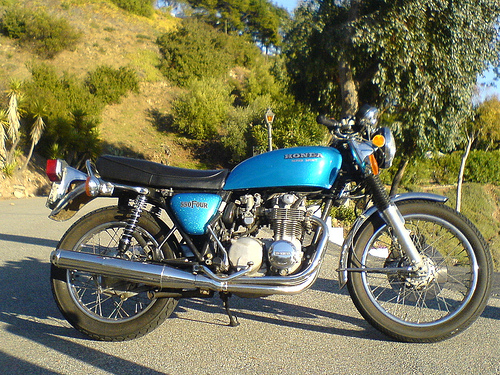} &
			\includegraphics[width=.2\linewidth]{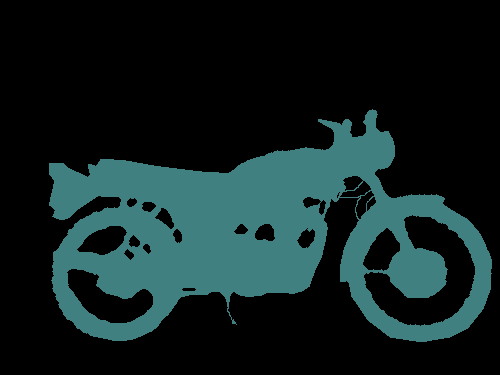} &
			\includegraphics[width=.2\linewidth]{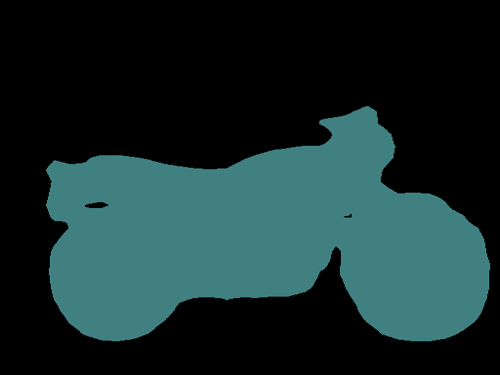} &
			\includegraphics[width=.2\linewidth]{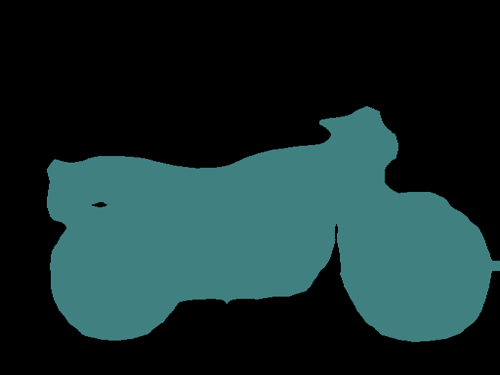} &
			\includegraphics[width=.2\linewidth]{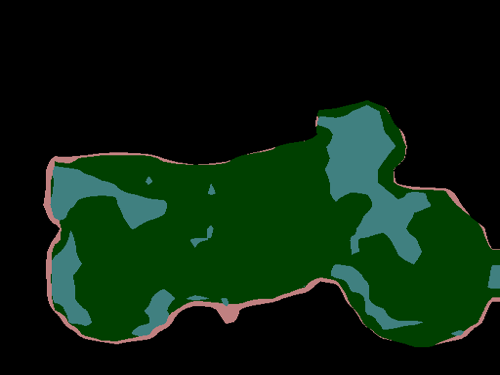} &
			\includegraphics[width=.2\linewidth]{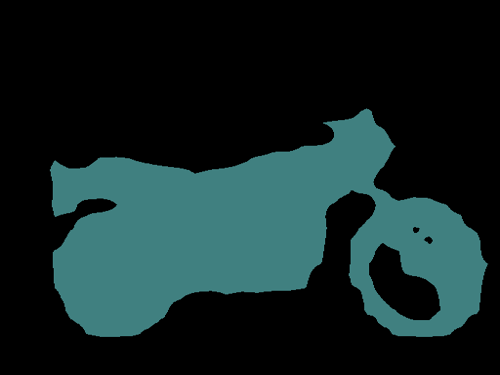} & \raisebox{1.6cm}{\rotatebox{-90}{\textbf{10-1}}} \\[-0.5ex]

		\end{tabular}
	\end{adjustbox}
	\caption{Qualitative visualizations from the PASCAL VOC \citep{PASCAL} dataset.}
	\label{fig:supp:voc}
	
\end{figure*}

\begin{figure*}[t]
	\begin{adjustbox}{width=\textwidth}
		\centering
		\setlength{\tabcolsep}{1pt}
		\begin{tabular}{ccccccc} 
			\small
			\renewcommand\arraystretch{0.1}
			\textbf{Image} & \textbf{GT} & \textbf{JT} & \textbf{CLoRA (JT)} & \textbf{MiB} & \textbf{CLoRA} &  \\ 
			
			\includegraphics[width=.2\linewidth]{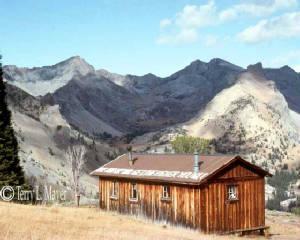} &
			\includegraphics[width=.2\linewidth]{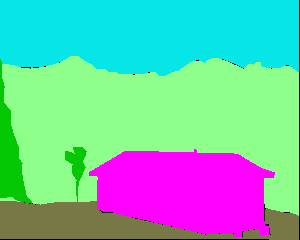} &
			\includegraphics[width=.2\linewidth]{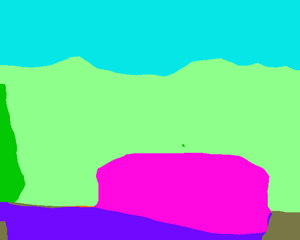} &
			\includegraphics[width=.2\linewidth]{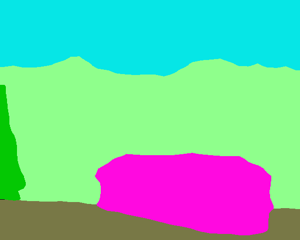} &
			\includegraphics[width=.2\linewidth]{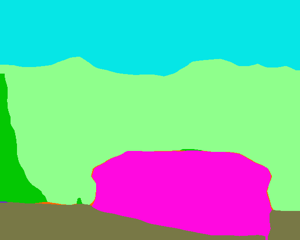} &
			\includegraphics[width=.2\linewidth]{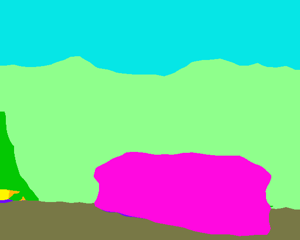} & \raisebox{1.55cm}{\rotatebox{-90}{\textbf{100-50}}} \\[-0.5ex]
			
			\includegraphics[width=.2\linewidth]{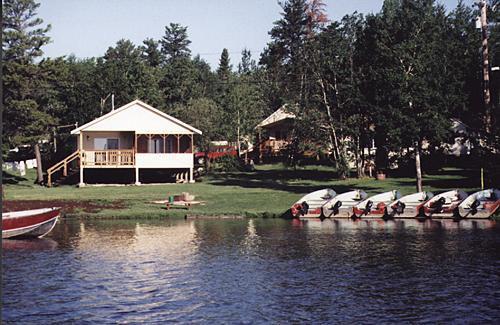} &
			\includegraphics[width=.2\linewidth]{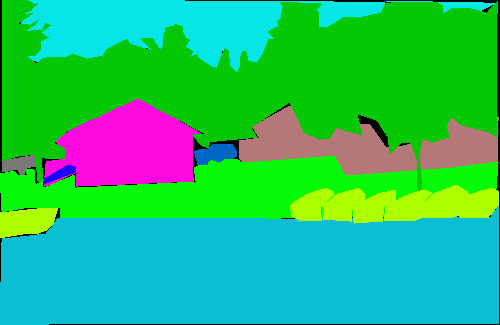} &
			\includegraphics[width=.2\linewidth]{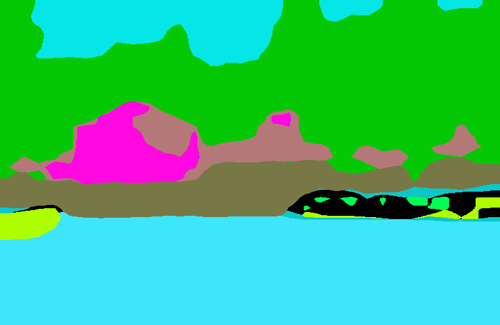} &
			\includegraphics[width=.2\linewidth]{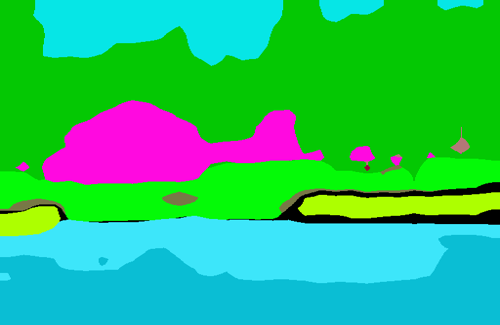} &
			\includegraphics[width=.2\linewidth]{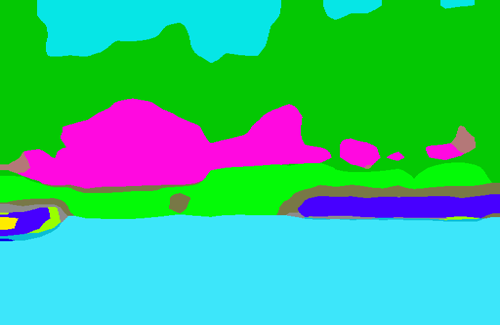} &
			\includegraphics[width=.2\linewidth]{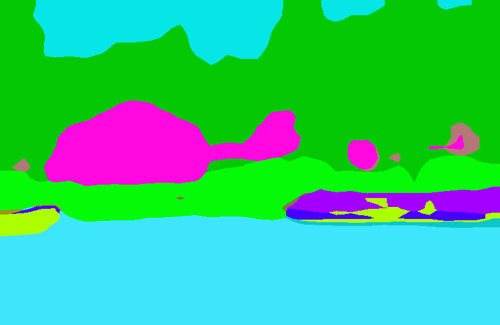} & \raisebox{1.45cm}{\rotatebox{-90}{\textbf{50-50}}} \\[-0.5ex]
			
			\includegraphics[width=.2\linewidth]{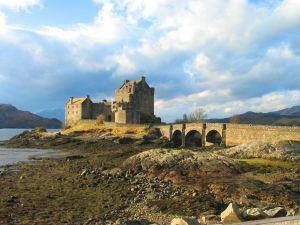} &
			\includegraphics[width=.2\linewidth]{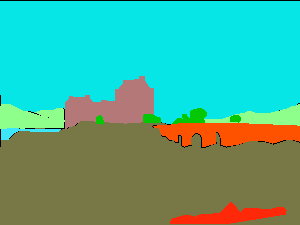} &
			\includegraphics[width=.2\linewidth]{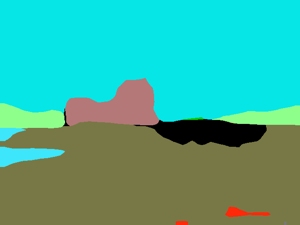} &
			\includegraphics[width=.2\linewidth]{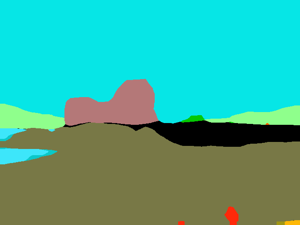} &
			\includegraphics[width=.2\linewidth]{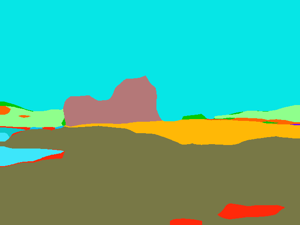} &
			\includegraphics[width=.2\linewidth]{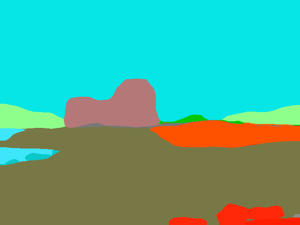} & \raisebox{1.65cm}{\rotatebox{-90}{\textbf{25-25}}} \\[-0.5ex]
			
			\includegraphics[width=.2\linewidth]{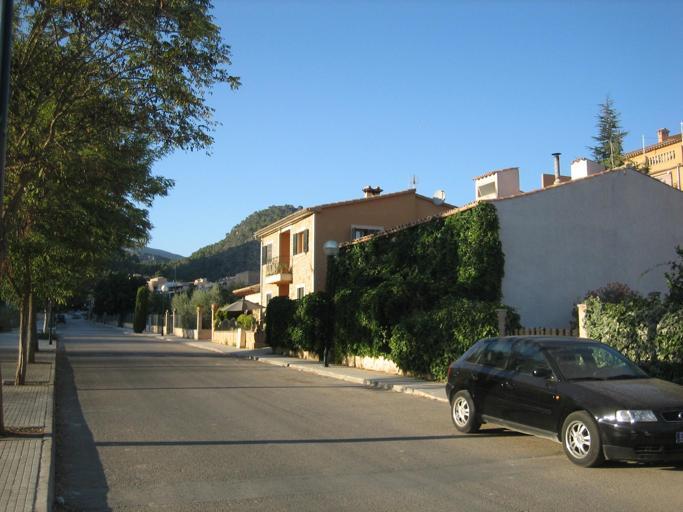} &
			\includegraphics[width=.2\linewidth]{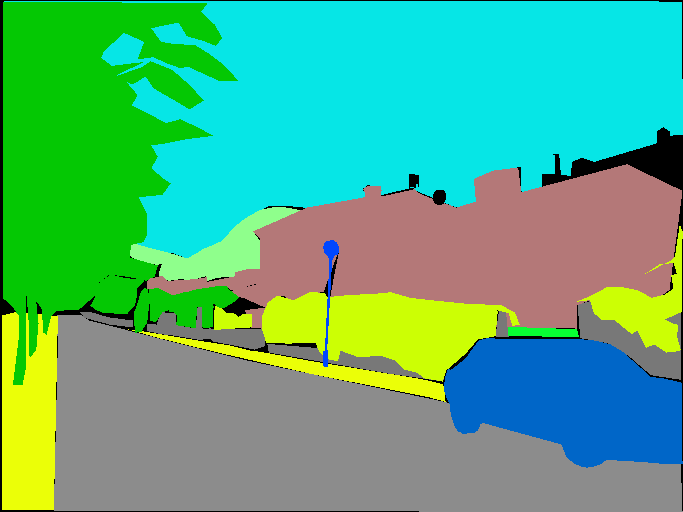} &
			\includegraphics[width=.2\linewidth]{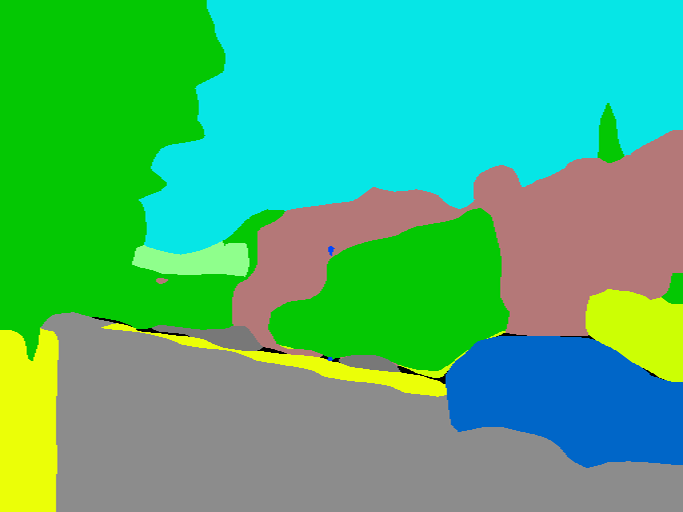} &
			\includegraphics[width=.2\linewidth]{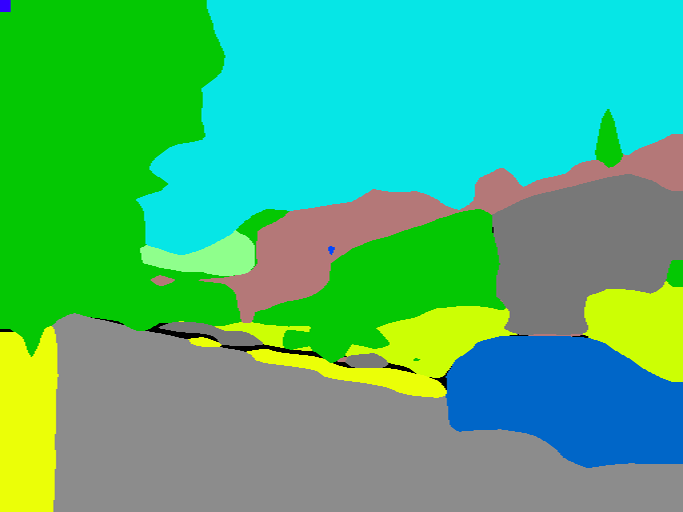} &
			\includegraphics[width=.2\linewidth]{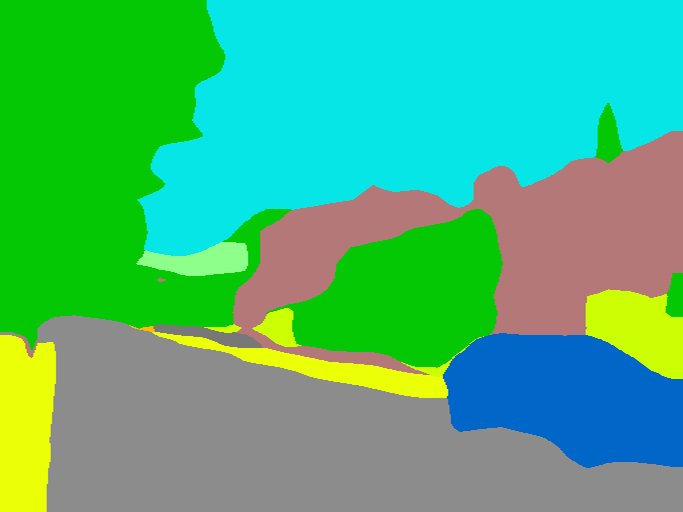} &
			\includegraphics[width=.2\linewidth]{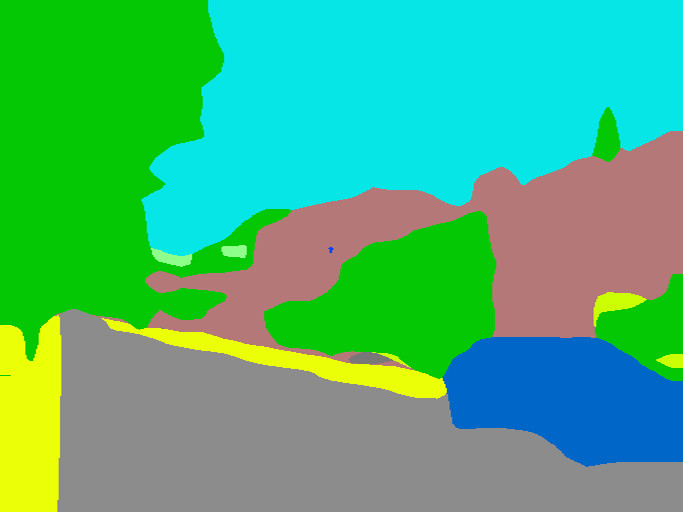} & \raisebox{1.45cm}{\rotatebox{-90}{\textbf{100-10}}} \\[-0.5ex]
			
			
		\end{tabular}
	\end{adjustbox}
	\caption{Qualitative visualizations from the ADE20K \citep{ADE} dataset.}
	\label{fig:supp:ade}
\end{figure*}

\Cref{fig:supp:voc} provides results of the \textit{15-5}, \textit{15-1}, \textit{5-3} and \textit{10-1}  experiments on PASCAL VOC \citep{PASCAL}.
The four experiments \textit{100-50}, \textit{50-50}, \textit{25-25}, and \textit{100-10} on ADE20K \citep{ADE} are visualized in \cref{fig:supp:ade}.
Despite its increased efficiency, CLoRA appears competitive or even more detailed, compared to previous work \citep{MiB}, in all four experiments.

\section{Effect of LoRA Rank}

\Cref{tab:sup:voc_ranks} presents the results for the PASCAL VOC \citep{PASCAL} tasks for different ranks of LoRA \citep{LoRA}. 
The rank is a hyperparameter that influences the number of trainable parameters. 
In all our main experiments with the ViT-based network, we use a rank $r=32$ which corresponds to $\sim$1\% of trainable parameters. 
In \cref{sec:experiments:additional:ranks} of the main paper, we study the influence of varying ranks in the offline setting (joint training) across the three datasets. 
Here, we explore the impact of different ranks on performance in more detail in different continual learning settings of the PASCAL VOC dataset \citep{PASCAL}. 
We observe that for rank $r=64$, the results improve consistently across all tasks. 
For longer task sequences, such as \textit{5-3} and \textit{10-1}, higher ranks yield better results. 
However, since task sequence lengths are typically unknown and higher ranks entail greater computational costs, we choose $r=32$ as a balanced configuration for all experiments.

\begin{table*}[t]
	\caption{Results of CLoRA on PASCAL VOC \citep{PASCAL} dataset with varying ranks for LoRA \citep{LoRA} after learning all tasks.}
	\begin{adjustbox}{width=\textwidth}
		\centering
		\begin{tabular}{c||c|c|c||c|c|c||c|c|c||c|c|c}
			\boldhline
			
			\multirow{2}{*}{\textbf{Rank}} & \multicolumn{3}{c||}{\textbf{15-5}} & \multicolumn{3}{c||}{\textbf{15-1}} & \multicolumn{3}{c||}{\textbf{5-3}} & \multicolumn{3}{c}{\textbf{10-1}} \\ \cline{2-13}
			&  0-15  &  16-20 &  \textbf{All}  &  0-15  &  16-20 &  \textbf{All}  &  0-5   &  6-20  &  \textbf{All}  &  0-10  &  11-20 &  \textbf{All}   \\ \boldhline
			
			
			16   & 74.87  & 55.82  & 70.34 &  80.71  & 34.37 & 69.67 & 67.06  & 47.32 & 52.96 & 28.05 & 26.23  & 27.19  \\     \hline
			
			32  & 74.17  & 56.57  & 70.39 & 81.29 & 34.41 & 70.13 & 69.92  & 45.50  & 52.47 & 31.38  & 29.22  & 30.35 \\ \hline
			
			64 & 79.62 & 62.31  & \textbf{75.50} & 82.06 & 34.08 & \textbf{70.63} & 71.46 & 45.70 & 53.06 & 47.62  & 33.71  & \textbf{41.00}  \\ \hline
			
			96 & 74.45 & 56.30  & 70.13 & 77.18 & 29.43 & 65.82 & 69.92 &  50.08 & \textbf{55.75} & 48.15  & 30.97  & 39.97  \\ \hline
			
			128  & 75.39  & 57.42  & 71.11 & 81.01 & 31.76  & 69.28  &  66.21 & 50.66 &  55.10 & 42.57 & 32.61  & 37.83    \\ \boldhline
		\end{tabular}
	\end{adjustbox}
	\label{tab:sup:voc_ranks}	
\end{table*}

\end{document}

%% file: math_commands.tex
\usepackage{amsmath,amsfonts,bm}

\def\eqref#1{equation~\ref{#1}}

\def\1{\bm{1}}

\def\vs{{\bm{s}}}

\DeclareMathAlphabet{\mathsfit}{\encodingdefault}{\sfdefault}{m}{sl}
\SetMathAlphabet{\mathsfit}{bold}{\encodingdefault}{\sfdefault}{bx}{n}

